\documentclass[runningheads]{llncs}
\usepackage{graphicx}

\usepackage{tikz}
\usepackage{comment}
\usepackage{amsmath,amssymb} 
\usepackage{color}

\usepackage[accsupp]{axessibility}  

\usepackage[width=122mm,left=12mm,paperwidth=146mm,height=193mm,top=12mm,paperheight=217mm]{geometry}

\usepackage{booktabs}
\usepackage{bm}
\usepackage{mathtools}
\usepackage{algpseudocode}
\usepackage{algorithm}
\usepackage{siunitx}
\usepackage{multirow}
\usepackage{enumitem}
\usepackage{xcolor}
\usepackage{diagbox}
\usepackage{subfigure}

\usepackage[pagebackref,breaklinks,colorlinks]{hyperref}

\makeatletter

\usepackage[capitalize]{cleveref}
\crefname{section}{Sec.}{Secs.}
\Crefname{section}{Section}{Sections}
\Crefname{table}{Table}{Tables}
\crefname{table}{Tab.}{Tabs.}

\begin{document}
\pagestyle{headings}
\mainmatter
\def\ECCVSubNumber{6207}  

\title{Network Binarization via Contrastive Learning} 

\graphicspath{ {images/} }

\algnewcommand\algorithmiclocalize{\textbf{Localize neuron and update model:}}
\algnewcommand\Localize{\item[\algorithmiclocalize]}
\algnewcommand\algorithmiccross{\textbf{Cross-section plane determination:}}
\algnewcommand\Cross{\item[\algorithmiccross]}
\algnewcommand\algorithmicbranchc{\textbf{Bifurcation candidates detection:}}
\algnewcommand\Branchc{\item[\algorithmicbranchc]}

\newcommand{\fft}[1]{\bm{\mathbf{\hat{#1}}}}
\newcommand{\norm}[1]{\vert \vert{#1}\vert \vert^2}
\newcommand{\normd}[1]{\vert \vert{#1}\vert \vert^2_2}
\newcommand{\diag}[1]{\text{diag}}
\newcommand{\conj}[1]{\text{conj}}

\newcommand{\syz}[1]{\textcolor{teal}{#1}}

\renewcommand{\vec}[1]{\boldsymbol{#1}}

\author{Yuzhang Shang\inst{1}\and
Dan Xu\inst{2} \and Ziliang Zong\inst{3}\and Liqiang Nie\inst{4}\and
Yan Yan\inst{1}\thanks{Corresponding author.}
}
\authorrunning{Y. Shang et al.}
%
\institute{Illinois Institute of Technology, USA \and
Hong Kong University of Science and Technology, Hong Kong \and
Texas State University, USA \and Harbin Institute of Technology, Shenzhen, China\\
 \email{yshang4@hawk.iit.edu, danxu@cse.ust.hk, ziliang@txstate.edu, nieliqiang@gmail.com, and yyan34@iit.edu}
}
\maketitle

\begin{abstract}
Neural network binarization accelerates deep models by quantizing their weights and activations into 1-bit. However, there is still a huge performance gap between Binary Neural Networks (BNNs) and their full-precision (FP) counterparts. As the quantization error caused by weights binarization has been reduced in earlier works, the activations binarization becomes the major obstacle for further improvement of the accuracy. BNN characterises a unique and interesting structure, where the binary and latent FP activations exist in the same forward pass (\textit{i.e.} $\text{Binarize}(\mathbf{a}_F) = \mathbf{a}_B$). To mitigate the information degradation caused by the binarization operation from FP to binary activations, we establish a contrastive learning framework while training BNNs through the lens of Mutual Information (MI) maximization. MI is introduced as the metric to measure the information shared between binary and the FP activations, which assists binarization with contrastive learning. Specifically, the representation ability of the BNNs is greatly strengthened via pulling the positive pairs with binary and FP activations from the same input samples, as well as pushing negative pairs from different samples (the number of negative pairs can be exponentially large). This benefits the downstream tasks, not only classification but also segmentation and depth estimation,~\textit{etc}. The experimental results show that our method can be implemented as a pile-up module on existing state-of-the-art binarization methods and can remarkably improve the performance over them on CIFAR-10/100 and ImageNet, in addition to the great generalization ability on NYUD-v2. The code is available at \url{https://github.com/42Shawn/CMIM}.

\end{abstract}
\section{Introduction}
\label{sec:intro}
Although deep learning~\cite{lecun2015deep} has achieved remarkable success in various computer vision tasks such as image classification~\cite{krizhevsky2012imagenet} and semantic image segmentation~\cite{chen2017deeplab}, its over-parametrization problem makes its computationally expensive and storage excessive. To advance the development of deep learning in resource-constrained scenarios, several neural network compression paradigms have been proposed, such as network pruning~\cite{lecun1989optimal,han2015deep}, knowledge distillation~\cite{hinton2015distilling,shang2021lipschitz} and network quantization~\cite{hubara2016binarized}. Among the network quantization methods, the network binarization method stands out for quantizing weights and activations (\textit{i.e.}~intermediate feature maps) to $\pm 1$, compressing the full-precision counterpart 32$\times$, and replacing time-consuming inner-product in full-precision networks with efficient xnor-bitcount operation in the BNNs~\cite{hubara2016binarized}.

However, severe accuracy drop-off always exists between full-precision models and their binary counterparts. To tackle this problem, previous works mainly focus on reducing the quantization error induced by weights binarization~\cite{rastegari2016xnor,lin2020rotated}, and elaborately approximating binarization function to alleviate the the gradient mismatch issue in the backward propagation~\cite{liu2020bi,qin2020forward}. Indeed, they achieve the SoTA performance. Yet narrowing down the quantization error and enhancing the gradient transmission reach their bottlenecks~\cite{cai2017deep,kim2021improving}, since the 1W32A (only quantizing the weights into 1-bit, remaining the activations 32-bit) models are capable of performing as well as the full-precision models~\cite{he2020proxybnn,lin2020rotated}, implying that the activations binarization becomes the main issue for further performance improvement.

\begin{figure}[!t]
\subfigure[Contrastive instance learning]{
    \begin{minipage}{0.46\textwidth}
    \includegraphics[width=0.95\textwidth]{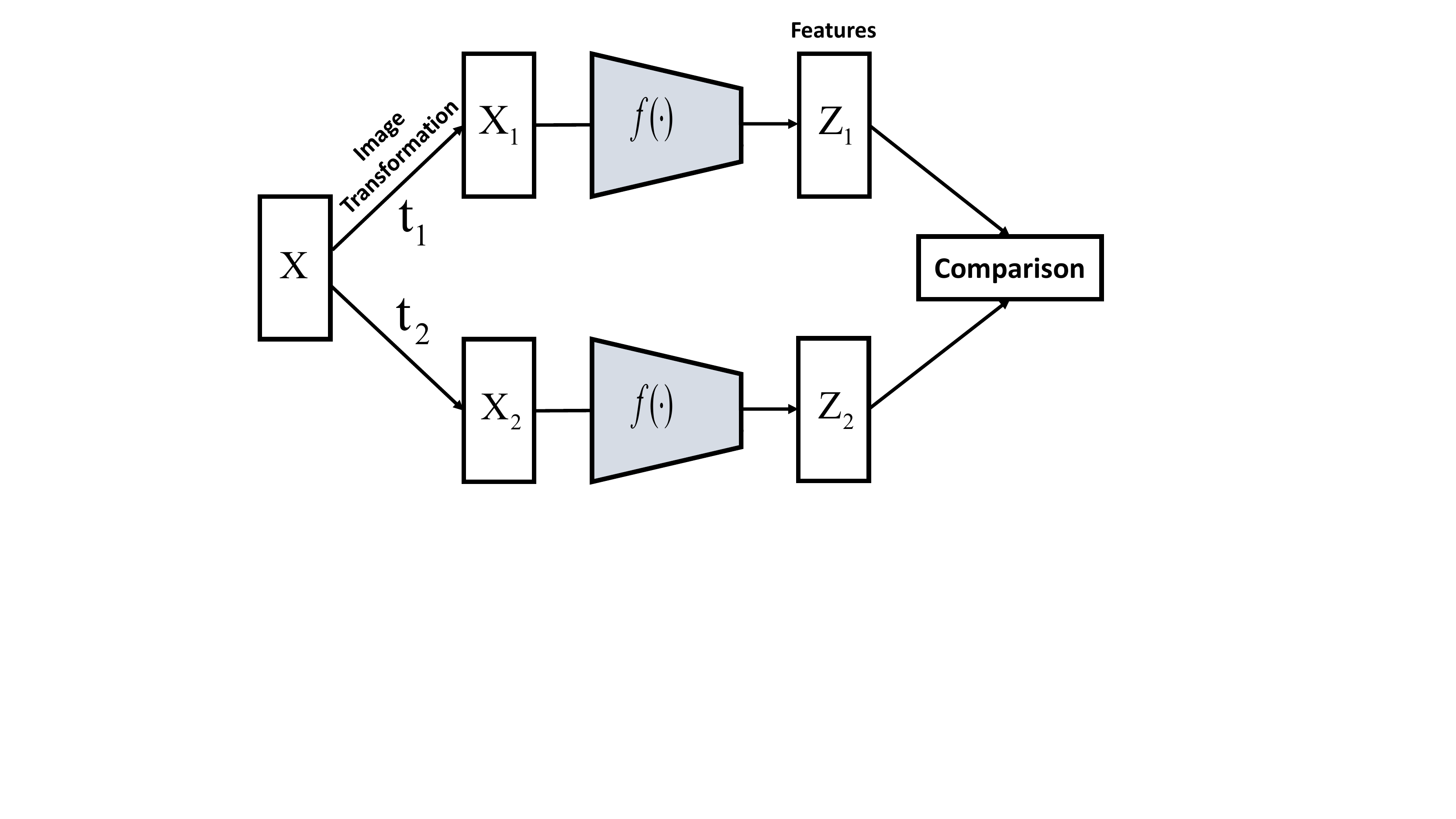}\vspace{0.05in}
    \end{minipage}}
    \label{pipeline:ctl}
\vspace{-0.1in}
\subfigure[\emph{CMIM} for training BNN]{
    \begin{minipage}{0.52\textwidth}
    \includegraphics[width=0.85\textwidth]{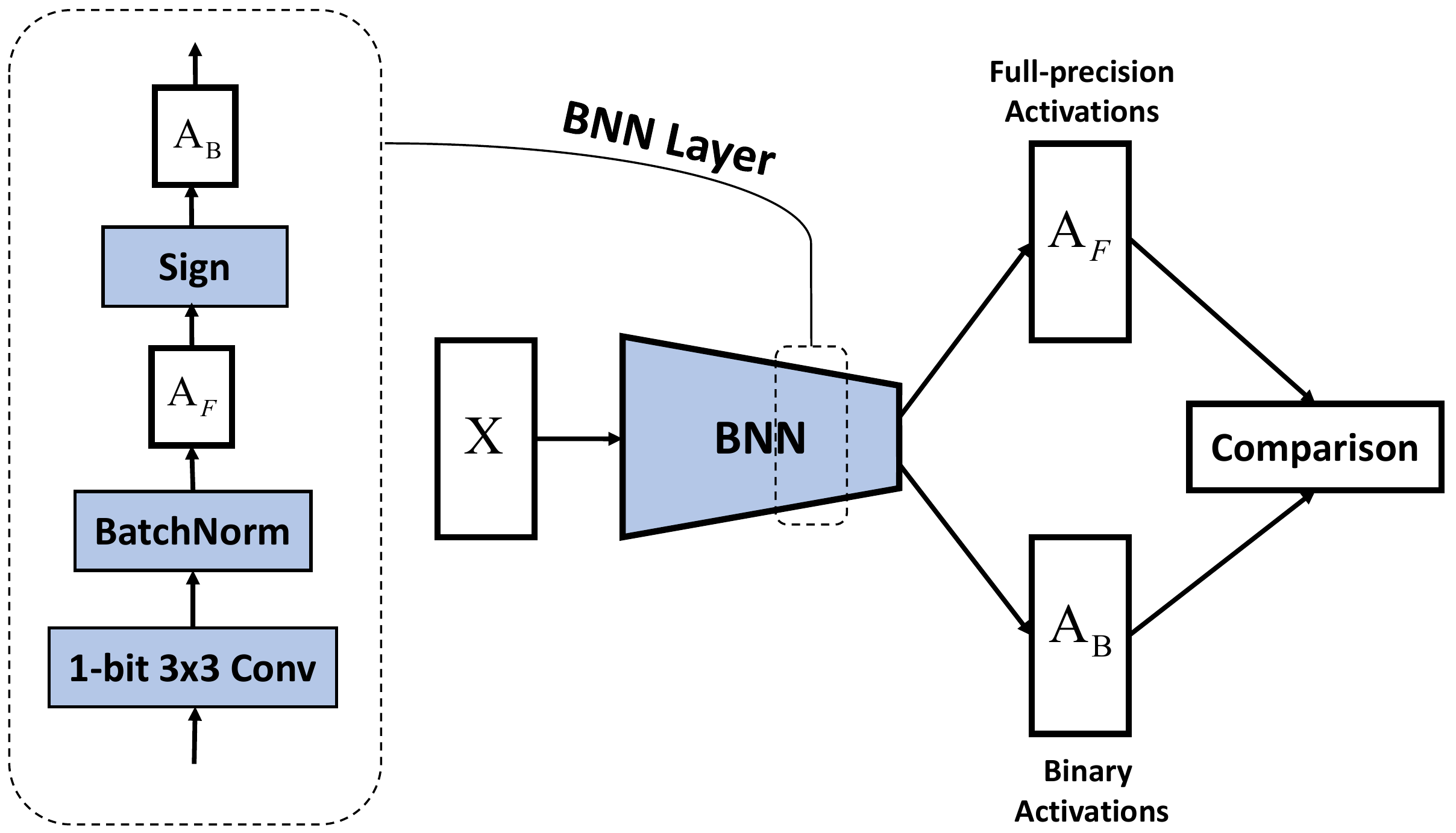}
    \end{minipage}}
    \label{fig:curve_c}
\caption{\textbf{(a)}: In contrastive instance learning, the features produced by different transformations of the same sample are contrasted to each other. \textbf{(b)}: However BNN can yield the binary activations $A_B$ and full-precision activations $A_F$ (\textit{i.e.} two transformations of an image both from the same BNN) in the same forward pass, thus the BNN can act as two image transformations in the literature of contrastive learning.}\label{fig:correlation}
\vspace{-0.8em}
\label{fig:pipeline}
\end{figure}

To address the accuracy degradation caused by the activations binarization, a few studies are proposed to regulate the distributions of the binary activations, \textit{e.g.} researchers in \cite{ding2019regularizing} design a distribution loss to explicitly regularize the activation flow; researchers in \cite{kim2021improving} propose to shift the thresholds of binary activation functions to unbalance the distribution of the binary activations. They heuristically design low-level patterns to analyze the distributions of binary activations, such as minimum of the activations and the balanced property of distributions. Nevertheless, they neglect the high-level indicators of the distribution and the unique characteristics of BNN, where the binary activations and latent full-precision activations co-exist in the same forward pass. Thus, we argue that the high-level properties of distributions, such as correlations and dependencies between binary and full-precision activations should be captured and utilized.

In this work, we explore introducing mutual information for BNNs, in which the mutual information acts as a metric to quantify the information amount shared by the binary and latent real-valued activations in BNNs. In contrast to the works mentioned above focusing on learning the distribution of binary activations, mutual information naturally captures statistical dependencies between variables, quantifying the degree of the dependence~\cite{gao2015efficient}. Based on this metric, we propose a novel method, termed as Network Binarization via \textbf{C}ontrastive Learning for \textbf{M}utual \textbf{I}nformation \textbf{M}aximization~(\textbf{\emph{CMIM}}). Specifically, we design a highly effective optimization strategy using contrastive estimation for mutual information maximization. As illustrated in Figure~\ref{fig:pipeline}, we replace the data transformation module in contrastive learning with the exclusive structure in BNNs, where full-precision and binary activations are in the same forward pass. In this way, contrastive learning contributes to inter-class decorrelation of binary activations, and avoids collapse solutions. In other words, our method is built upon a contrastive learning framework to learn representative binary activations, in which we pull the binary activation closer to the full-precision activation and push the binary activation further away from other binary activations in the contrastive space. Moreover, by utilizing an additional MLP module to extract representations of activations, our method can explicitly capture higher-order dependencies in the contrastive space. To the best of our knowledge, it is the first work aiming at maximizing the mutual information of the activations in BNNs within a contrastive learning framework.

Overall, the contributions of this paper are three-fold:
\begin{itemize}[leftmargin=*, topsep=0pt, partopsep=0pt, itemsep=1pt]
    \item Considering the distributions of activations, we propose a novel contrastive framework to optimize BNNs, via maximizing the mutual information between the binary activation and its latent real-valued counterpart;
    \item We develop an effective contrastive learning strategy to achieve the goal of mutual information maximization for BNNs, and benefited from it, the representation ability of BNNs is strengthened for not only the classification task but also downstream CV tasks;
    \item Experimental results show that our method can significantly improve the existing SoTA methods over the classification task on CIFAR-10/100 and ImageNet, \textit{e.g.} 6.4\% on CIFAR-100 and 3.0\% on ImageNet. Besides, we also demonstrate the great generalization ability of the proposed \emph{CMIM} on other challenging CV tasks such as depth estimation and semantic segmentation. 
\end{itemize}

\section{Related Work}
\label{related}
In \cite{hubara2016binarized}, the researchers introduce the sign function to binarize weights and activations to 1-bit, initiating the studies of BNNs. In this work, the straight-through estimator (STE)~\cite{bengio2013estimating} is utilized to approximate the derivative of the sign function. Following the seminal art, copious studies contribute to improving the performance of BNNs. For example, Rastegari \textit{et al.}~\cite{rastegari2016xnor} disclose that the quantization error between the full-precision weights and the corresponding binarized weights is one of the major obstacles degrading the representation capabilities of BNNs. Reducing the quantization error thus becomes a fundamental research direction to improve the performance of BNNs. Researchers propose XNOR-Net~\cite{rastegari2016xnor} to introduce a scaling factor calculated by L1 norm for both weights and activation functions to minimize the quantization error. Inspired by XNOR-Net, XNOR++~\cite{bulat2019xnor} further learns both spatial and channel-wise scaling factors to improves the performances. Bi-Real~\cite{liu2020bi} proposes double residual connections with full-precision downsampling layers to mitigate the excessive gradient vanishing issue caused by binarization. ProxyBNN~\cite{he2020proxybnn} designs a proxy matrix as a basis of the latent parameter space to guide the alignment of the weights with different bits by recovering the smoothness of BNNs. ReActNet~\cite{liu2020reactnet} implements binarization with MobileNet~\cite{howard2017mobilenets} instead of ResNet, and achieves SoTA performance.

Nevertheless, we argue that those methods focusing on narrowing down the quantization error and enhancing the gradient transmission reach their bottleneck (\textit{e.g.} 1W32A ResNet-18 trained by ProxyBNN achieves 67.7\% Top-1 accuracy on ImageNet, while full-precision version is only 68.5\%). Because they neglect the activations in BNNs, especially the relationship between the binary and latent full-precision activations. We treat them as discrete variables and investigate them under the metric of mutual information. By maximizing the mutual information via contrastive learning, the performance of BNNs is further improved. The experimental results show that \emph{CMIM} can consistently improve the aforementioned methods by directly adding our \emph{CMIM} module on them.

\section{Training BNNs via Contrastive Learning for Mutual Information Maximization}
\subsection{Preliminaries}
\label{sec:pre}
We define a $K$-layer Multi-Layer Perceptron (MLP). For simplification, we discard the bias term of this MLP. Then the network $f(\mathbf{x})$ can be denoted as:
\begin{equation}
    f(\mathbf{W}^1,\cdots,\mathbf{W}^K;\mathbf{x}) = (\mathbf{W}^{K}\cdot\sigma\cdot \mathbf{W}^{K-1}\cdot \cdots \cdot\sigma\cdot \mathbf{W}^{1})(\mathbf{x}),
    \label{eq:mlp1}
\end{equation}
where $\mathbf{x}$ is the input sample and $\mathbf{W}^{k}:\mathbb{R}^{d_{k-1}} \longmapsto \mathbb{R}^{d_{k}}(k=1,...,K)$ stands for the weight matrix connecting the $(k-1)$-th and the $k$-th layer, with $d_{k-1}$ and $d_{k}$ representing the sizes of the input and output of the $k$-th network layer, respectively. The $\sigma(\cdot)$ function performs element-wise activation operation on the input feature maps. 

Based on those predefined notions, the sectional MLP $f^k(\mathbf{x})$ with the front $k$ layers of the $f(\mathbf{x})$ can be represented as:
\begin{equation}
    f^k(\mathbf{W}^1,\cdots,\mathbf{W}^k;\mathbf{x}) = (\mathbf{W}^{k}\cdot\sigma\cdots\sigma\cdot \mathbf{W}^{1})(\mathbf{x}).
    \label{eq:mlp2}
\end{equation}
And the MLP $f$ can be seen as a special case in the function sequence $\{f^k\}(k\in\{1,\cdots,K\})$, \textit{i.e.} $f = f^K$.

\sloppy
\noindent \textbf{Binary Neural Networks.} Here, we review the general binarization method in~\cite{courbariaux2015binaryconnect,hubara2016binarized}, which maintains latent full-precision weights $\{\mathbf{W}_F^k\}(k\in\{1,\cdots,K\})$ for gradient updates, and the $k$-th weight matrix $\mathbf{W}_F^k$ is binarized into $\pm 1$, obtaining the binary weight matrix $\mathbf{W}_B^k$ by a binarize function (normally $\textit{sgn}(\cdot)$), \textit{i.e.} $\mathbf{W}_B^k = \textit{sgn}(\mathbf{W}_F^k)$. Then the intermediate activation map (full-precision) of the $k$-th layer is produced by $\mathbf{A}_F^{k} = \mathbf{W}_B^k \mathbf{A}_B^{k-1}$. Finally, the same sign function is used to binarize the full-precision activations into binary activations as $\mathbf{A}_B^k = \textit{sgn}(\mathbf{A}_F^k)$~(see~ Fig.\ref{pipeline:ctl} b), and the whole forward pass of a BNN is performed by iterating this process for $L$ times. 

\noindent\textbf{Mutual Information and Contrastive Learning.}
For two discrete variables $\mathbf{X}$ and $\mathbf{Y}$, their mutual information (MI) can be defined as~\cite{kullback1997information}:
\begin{equation}
    I(\mathbf{X}, \mathbf{Y}) = \sum_{x,y}P_{\mathbf{X}\mathbf{Y}}(x, y)\log\frac{P_{\mathbf{X}\mathbf{Y}}(x, y)}{P_{\mathbf{X}}(x)P_{\mathbf{Y}}(y)},
\label{eq:mi}
\end{equation}
where $P_{\mathbf{X}\mathbf{Y}}(x, y)$ is the joint distribution, $P_{\mathbf{X}}(x) = \sum_{y}P_{\mathbf{X}\mathbf{Y}}(x, y)$ and $P_{\mathbf{Y}}(y) = \sum_{x}P_{\mathbf{X}\mathbf{Y}}(x, y)$ are the marginals of $\mathbf{X}$ and $\mathbf{Y}$, respectively.

Mutual information quantifies the amount of information obtained about one random variable by observing the other random variable. It is a dimensionless quantity with (generally) units of bits, and can be considered as the reduction in uncertainty about one random variable given knowledge of another. High mutual information indicates a large reduction in uncertainty and \textit{vice versa}~\cite{kullback1997information}. In the content of binarization, considering the binary and full-precision activations as random variables, we would like them share as much information as possible, since the binary activations are proceeded from their corresponding full-precision activations. Theoretically, the mutual information between those two variables should be maximized. 

Our motivation can also be testified from the perspective of RBNN~\cite{lin2020rotated}. In RBNN, Lin \textit{et al.} devise a rotation mechanism leading to around 50\% weight flips which maximizes the information gain, $H(\mathbf{a}_B^{k,i})$. As MI can be written in another form as $I(\mathbf{X},\mathbf{Y}) = H(\mathbf{X}) - I(\mathbf{X} \mid \mathbf{Y})$, the MI between binary and FP activations can be formulated as:
\begin{equation}
    I(\mathbf{a}_B^{k,i},\mathbf{a}_F^{k,j}) = H(\mathbf{a}_B^{k,i}) - I(\mathbf{a}_B^{k,i} \mid \mathbf{a}_F^{k,j}),
    \label{eq:mi1}
\end{equation}
in which maximizing the first term on the right can partially lead to maximizing the whole MI. In this work, we aim to universally maximize the targeted MI.

\sloppy
Recently, contrastive learning is proven to be an effective approach to MI maximization, and many methods based on contrastive loss for self-supervised learning are proposed, such as Deep InfoMax~\cite{hjelm2018learning}, Contrastive Predictive Coding~\cite{oord2018representation}, MemoryBank~\cite{wu2018unsupervised}, Augmented Multiscale DIM~\cite{bachman2019learning}, MoCo~\cite{he2020momentum} and SimSaim~\cite{chen2021exploring}. These methods are generally rooted in NCE~\cite{gutmann2010noise} and InfoNCE~\cite{hjelm2018learning} which can serve as optimizing the lower bound of mutual information~\cite{poole2019variational}. 
Intuitively, the key idea of contrastive learning is to pull representations in positive pairs close and push representations in negative pairs apart in a contrastive space, and thus the major obstacle for resorting to the contrastive loss is to define the negative and positive pairs.

\subsection{Contrastive Learning for Mutual Information Maximization}
\label{sec:method}
In this section, we formalize the idea of constructing a contrastive loss based on Noise-Contrastive Estimation (NCE) to maximize the mutual information between the binary and the full-precision activations. Particularly, we derive a novel \emph{CMIM} loss for training BNNs, where NCE is introduced to avoid the direct mutual information computation by estimating it with its lower bound in Eq.~\ref{eq:ctl4}. Straightforwardly, the binary and full-precision activations from samples can be pull close, and activations from different samples can be pushed away, which corresponds to the core idea of contrastive learning.

For binary network $f_B$ and its latent full-precision counterpart $f_F$ in the same training iteration, the series of their activations $\{\mathbf{a}_B^k\}$ and $\{\mathbf{a}_F^k\}(k\in\{1,\cdots,K\})$, where $\mathbf{A}_B^k = (\mathbf{a}_B^{k,1},\cdots,\mathbf{a}_B^{k,N})$ and $\mathbf{A}_F^k = (\mathbf{a}_F^{k,1},\cdots,\mathbf{a}_F^{k,N})$ can be considered as a series of variables. The corresponding variables $(\mathbf{a}_B^k,\mathbf{a}_F^k)$ should share more information, \textit{i.e.} the mutual information of the same layer's output activations $I(\mathbf{a}_B^k, \mathbf{a}_F^k)(k\in\{1,\cdots,K\})$ should be maximized to enforce them mutually dependent.

\begin{figure}[!t]
\centering
  \includegraphics[width=0.6\textwidth]{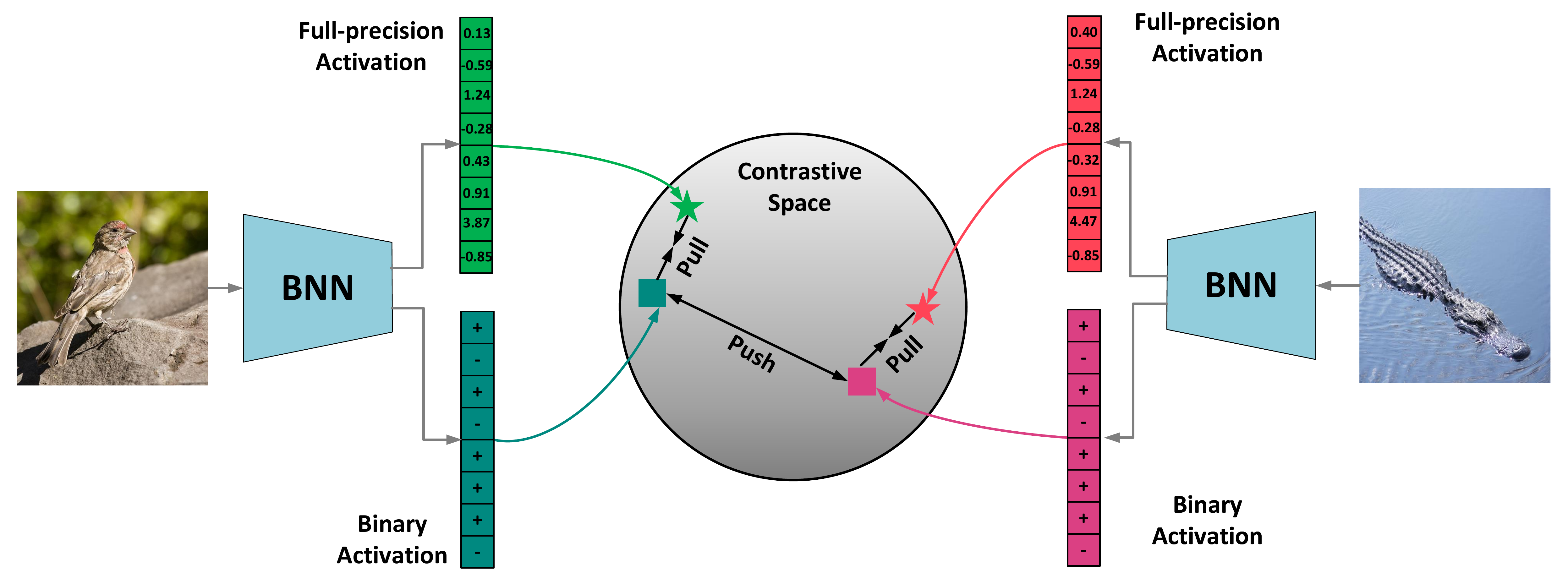}
  \caption{Feeding two images into a BNN, and obtaining the three pairs of binary and full-precision activations. Our goal is to embed the activations into a contrastive space, then learn from the pair correlation with the contrastive learning task in Eq.~\ref{eq:loss1}.}
  \vspace{-0.15in}
  \label{fig:intuition}
\end{figure}

To this end, we introduce the contrastive learning framework into our targeted binarization task. The basic idea of contrastive learning is to compare different views of the data (usually under different data augmentations) to calculate similarity scores~\cite{oord2018representation,hjelm2018learning,bachman2019learning,he2020momentum,chen2021exploring}. This framework is suitable for our case, since the binary and full-precision activations can be seen as two different views. For a training batch with $N$ samples, the samples can be denoted as: $\{\mathbf{x}_i\}(i\in\{1,\cdots,N\})$. We feed a batch of samples to the BNN and obtain $KN^2$ pairs of activations $(\mathbf{a}_B^{k,i},\mathbf{a}_F^{k,j})$, which augments the data for the auxiliary task. We define a pair containing two activations from the same sample as positive pair, \textit{i.e.} if $i=j$, $(\mathbf{a}_B^{k,i},\mathbf{a}_F^{k,j})_{+}$ and \textit{vice versa}. The core idea of contrastive learning is to discriminate whether a given pair of activation $( \mathbf{a}_B^{k,i},\mathbf{a}_F^{k,j})$ is positive or negative, \textit{i.e.}, inferring the distribution $P(D \mid \mathbf{a}_B^{k,i},\mathbf{a}_F^{k,j})$, in which $D$ is the variable decides whether $i = j$ or $i \neq j$. However, we can not directly compute the distribution $P(D \mid \mathbf{a}_B^{k,i},\mathbf{a}_F^{k,j})$~\cite{gutmann2010noise}, and we introduce its variational approximation
\begin{equation}
    q(D \mid \mathbf{a}_B^{k,i},\mathbf{a}_F^{k,j}),
\label{qu:ctl0}
\end{equation}
which can be calculated by our models. Intuitively, $q(D \mid \mathbf{a}_B^{k,i},\mathbf{a}_F^{k,j})$ can be treated as a binary classifier, which can classify a given pair $(\mathbf{a}_B^{k,i},\mathbf{a}_F^{k,j})$ into positive or negative.

With the Bayes' theorem, the posterior probability of two activations from the positive pair can be formalized as: 
\begin{equation}
    q(D = 1 \mid \mathbf{a}_B^{k,i},\mathbf{a}_F^{k,j}) = \frac{q(\mathbf{a}_B^{k,i},\mathbf{a}_F^{k,j}\mid D=1)\frac{1}{N}}{q(\mathbf{a}_B^{k,i},\mathbf{a}_F^{k,j}\mid D = 1)\frac{1}{N} + q(\mathbf{a}_B^{k,i},\mathbf{a}_F^{k,j}\mid D = 1)\frac{N-1}{N}}.
\label{eq:ctl1}
\end{equation}

The probability of activations from negative pair is $q(D = 0\mid \mathbf{a}_B^{k,i},\mathbf{a}_F^{k,j}) = 1 - q(D = 1\mid \mathbf{a}_B^{k,i},\mathbf{a}_F^{k,j})$.
To simplify the NCE derivative, several works~\cite{gutmann2010noise,wu2018unsupervised,tian2019contrastive} build assumption about the dependence of the variables, we also use the assumption that the activations from positive pairs are dependent and the ones from negative pairs are independent, \textit{i.e.} $q(\mathbf{a}_B^{k,i},\mathbf{a}_F^{k,j}\mid D=1) = P(\mathbf{a}_B^{k,i},\mathbf{a}_F^{k,j})$ and $q(\mathbf{a}_B^{k,i},\mathbf{a}_F^{k,j}\mid D=0) = P(\mathbf{a}_B^{k,i})P(\mathbf{a}_F^{k,j})$.
Hence, the above equation can be simplified as:
\begin{equation}
    q(D = 1 \mid \mathbf{a}_B^{k,i},\mathbf{a}_F^{k,j}) = \frac{P(\mathbf{a}_B^{k,i},\mathbf{a}_F^{k,j})}{P(\mathbf{a}_B^{k,i},\mathbf{a}_F^{k,j}) + P(\mathbf{a}_B^{k,i})P(\mathbf{a}_F^{k,j})(N-1)}.
\label{eq:ctl2}
\end{equation}
Performing logarithm to Eq.~\ref{eq:ctl2} and arranging the terms, we can achieve
\begin{equation}
\begin{aligned}
 \log q(D = 1 \mid \mathbf{a}_B^{k,i},\mathbf{a}_F^{k,j}) \leq \log\frac{P(\mathbf{a}_B^{k,i},\mathbf{a}_F^{k,j})}{P(\mathbf{a}_B^{k,i})P(\mathbf{a}_F^{k,j})} - \log(N-1).
\label{eq:ctl3}
\end{aligned}
\end{equation}

Taking expectation on both sides with respect to $P(\mathbf{a}_B^{k,i},\mathbf{a}_F^{k,j})$, and combining the definition of mutual information in Eq.~\ref{eq:mi}, we can derive the form of mutual information as: 
\begin{equation}
\begin{aligned}
    \overbrace{I(\mathbf{a}_B^{k},\mathbf{a}_F^{k})}^{\text{targeted \textbf{MI}}}  
    \geq \overbrace{\mathbb{E}_{P(\mathbf{a}_B^{k,i},\mathbf{a}_F^{k,j})}\left[\log q(D=1\mid \mathbf{a}_B^{k,i},\mathbf{a}_F^{k,j})\right]}^{\text{ optimized lower bound}} + \log(N-1),
\label{eq:ctl4}
\end{aligned}
\end{equation}
where $I(\mathbf{a}_B^{k},\mathbf{a}_F^{k})$ is the mutual information between the binary and full-precision distributions of our targeted object. Instead of directly maximizing the mutual information, maximizing the lower bound in the Eq.~\ref{eq:ctl4} is a practical solution. 

However, $q(D =1 \mid \mathbf{a}_B^{k,i},\mathbf{a}_F^{k,j})$ is still hard to estimate. Thus, we introduce critic function $h$ with parameter $\phi$ (\textit{i.e.} $h(\mathbf{a}_B^{k,i},\mathbf{a}_F^{k,j};\phi)$) as previous contrastive learning works~\cite{oord2018representation,hjelm2018learning,bachman2019learning,tian2019contrastive,chen2021wasserstein}. Basically, the critic function $h$ needs to map $\mathbf{a}_B^{k},\mathbf{a}_F^{k}$ to $\left[0, 1\right]$ (\textit{i.e.} discriminate whether a given pair is positive or negative). In practice, we design our critic function for our BNN case based on the critic function in~\cite{tian2019contrastive}:
\begin{equation}
    h(\mathbf{a}_B^{k,i},\mathbf{a}_F^{k,j}) =\exp(\frac{<\mathbf{a}_B^{k,i}, \mathbf{a}_F^{k,j}>}{\tau})/C,
    \label{eq:critic}
\end{equation}
in which $C = \exp(\frac{<\mathbf{a}_B^{k,i}, \mathbf{a}_F^{k,j}>}{\tau}) + N/M$, $M$ is the number of all possible pairs, as well as $\tau$ is a temperature parameter that controls the concentration level of the distribution~\cite{hinton2015distilling}. 

The activations of BNN have their properties can be used here, \textit{i.e.}
\begin{equation}
    \textit{sgn}(\mathbf{a}_F^{k,i}) = \mathbf{a}_B^{k,i}~~~\text{and}~~~<\mathbf{a}_B^{k,i},\mathbf{a}_F^{k,i}> = \Vert \mathbf{a}_F^{k,i} \Vert_1
    \label{eq:critic_1}
\end{equation}
Thus, the critic function in Eq.~\ref{eq:critic} can be further simplified as follows:
\begin{equation}
h(\mathbf{a}_B^{k,i},\mathbf{a}_F^{k,j}) = \exp(\frac{<\textit{sgn}(\mathbf{a}_F^{k,i}), \mathbf{a}_F^{k,j}>}{\tau})= \left\{
    \begin{array}{ll}
        \exp(\frac{\Vert \mathbf{a}_F^{k,i} \Vert_1}{\tau}) & \quad i = j, \\
        \exp(\frac{<\textit{sgn}(\mathbf{a}_F^{k,i}), \mathbf{a}_F^{k,j}>}{\tau}) & \quad i \neq j  
    \end{array}
    \right.
    \label{eq:critic_new}
\end{equation}
\noindent\textbf{Critic in the view of activation flip.} Eq.~\ref{eq:critic_new} reveals the working mechanism of \emph{CMIM} from a perspective of activation flip. Specifically, by turning the $+$ activation into $-$, binary activation in the critic can pull the activations in positive pair close and push the ones in the negative pair away via inner product. For example, suppose $\mathbf{a}_F^{k,1}{=}(0.3,-0.4,-0.6)$ and $\mathbf{a}_F^{k,2}{=}(0.6,-0.9, 0.7)$, and then $\mathbf{a}_B^{k,1}{=}(+1,-1,-1)$ is the anchor. Thus, for the positive pair, $<\textit{sgn}(\mathbf{a}_B^{k,1}), \mathbf{a}_F^{k,1}>{=}0.3\times(+1)+(-0.4)\times(-1)+(-0.6)\times(-1){=}\Vert \mathbf{a}_F^{k,1}\Vert_1$ maximizing their similarity score; and for the negative pair, $<\textit{sgn}(\mathbf{a}_B^{k,1}),\mathbf{a}_F^{k,2}>{=}0.6\times(+1)+(-0.9)\times(-1)+\overbrace{(0.6)\times(-1)}^{flipped}$ gradually minimizing the score, where the flipped term serve as a penalty for the negative pair. In this way, the binary anchor pull the positive full-precision activation close, and push the negative full-precision ones away by flipping numbers in the full-precision activations. Note that the process is iteratively operated during training, and thus all the binary activations can play the role as anchor, which eventually leads to better representation capacity in the contrastive space.

\noindent\textbf{Loss Function.} We define the contrastive loss function $\mathcal{L}^{k}_{NCE}$ between the $k$-th layer's activations $\mathbf{A}_B^{k}$ and $\mathbf{A}_F^{k}$ as: $\mathcal{L}^{k}_{NCE} =$

\begin{equation}
      \mathbb{E}_{q(\mathbf{a}_B^{k,i},\mathbf{a}_F^{k,j}\mid D=1)}\left[\log h(\mathbf{a}_B^{k,i},\mathbf{a}_F^{k,j})\right] + N\mathbb{E}_{q(\mathbf{a}_B^{k,i},\mathbf{a}_F^{k,j}\mid D=0)}\left[\log(1 -  h(\mathbf{a}_B^{k,i},\mathbf{a}_F^{k,j}))\right].
    \label{eq:loss1}
\end{equation}
We would comment on the above loss function from the perspective of contrastive learning. The first term of positive pairs is optimized for capturing more intra-class correlations and the second term of negative pairs is for inter-class decorrelation. Because the pair construction is instance-wise, the number of negative samples theoretically can be the size of the entire training set, \textit{e.g.} 1.2 million for ImageNet. With those additional hand-craft designed contrastive pairs for the proxy optimization problem in Eq.~\ref{eq:loss1}, the representation capacity of BNNs can be further improved, as many contrastive learning methods demonstrated~\cite{chen2021exploring,oord2018representation,hjelm2018learning,bachman2019learning}.


Combining the series of NCE loss from different layers $\left\{\mathcal{L}^{k}_{NCE}\right\}, (k=1,\cdots,K)$, the overall loss $\mathcal{L}$ can be defined as:
\begin{equation}
    \mathcal{L} = \lambda\sum_{k=1}^{K}\frac{\mathcal{L}^{k}_{NCE}}{\beta^{K-1-k}} + \mathcal{L}_{cls},
    \label{eq:loss2}
\end{equation}
where $\mathcal{L}_{cls}$ is the classification loss respect to the ground truth, $\lambda$ is used to control the degree of NCE loss, $\beta$ is a coefficient greater than $1$, and we denote the \emph{CMIM} loss as $\mathcal{L}_{CMIM} = \sum_{k=1}^{K}\frac{\mathcal{L}^{k}_{NCE}}{\beta^{K-1-k}}$. Hence, the ${\beta^{K-1-k}}$ decreases with $k$ increasing and consequently the $\frac{\mathcal{L}^{k}_{NCE}}{\beta^{K-1-k}}$ increases. In this way, the activations of latter layer can be substantially retained, which leads to better performance in practice. The complete training process of \emph{CMIM} is presented in \textbf{Algorithm} 1 in the Supplemental Materials.

\subsection{Discussion on CMIM}
\begin{figure}[!t]
\subfigure[XNOR~\cite{rastegari2016xnor}]{
    \begin{minipage}{0.22\textwidth}
    \includegraphics[width=\textwidth]{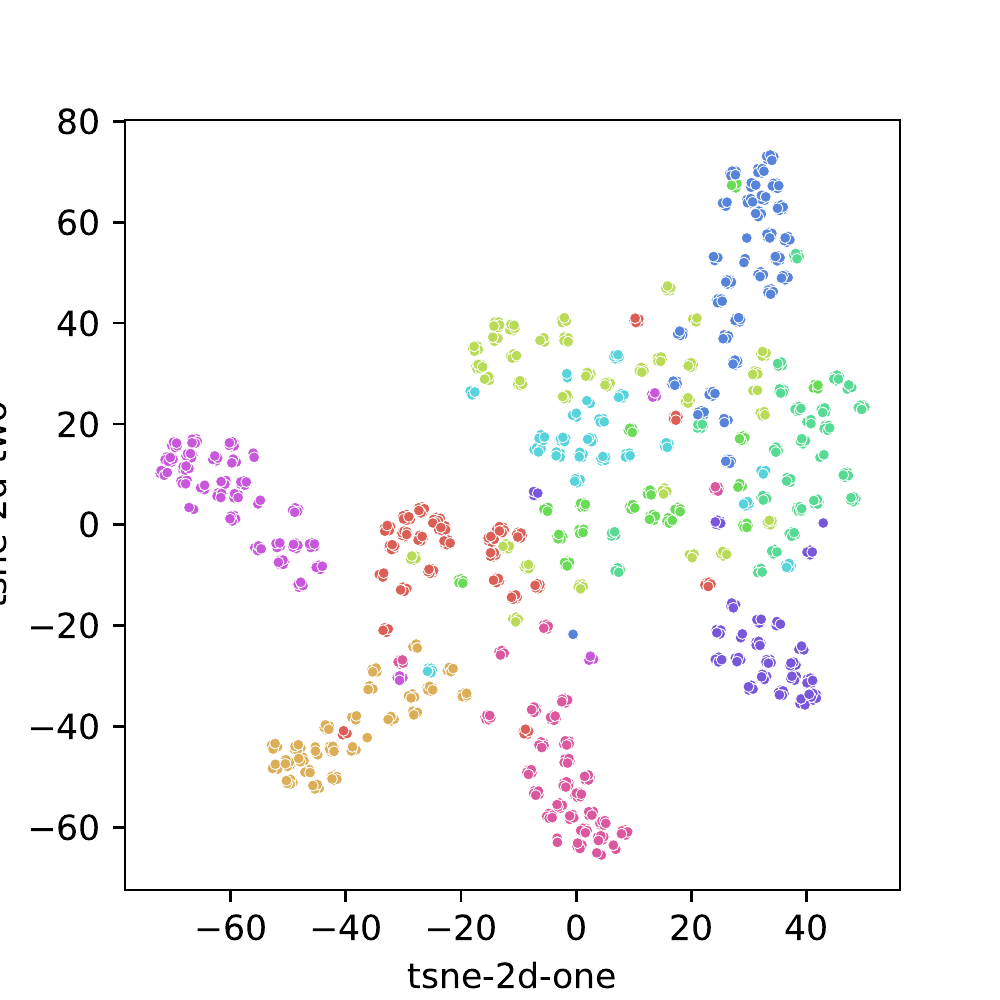}
    \end{minipage}}
    \label{fig:curve_a}
\subfigure[IR-Net~\cite{qin2020forward}]{
    \begin{minipage}{0.22\textwidth}
    \includegraphics[width=\textwidth]{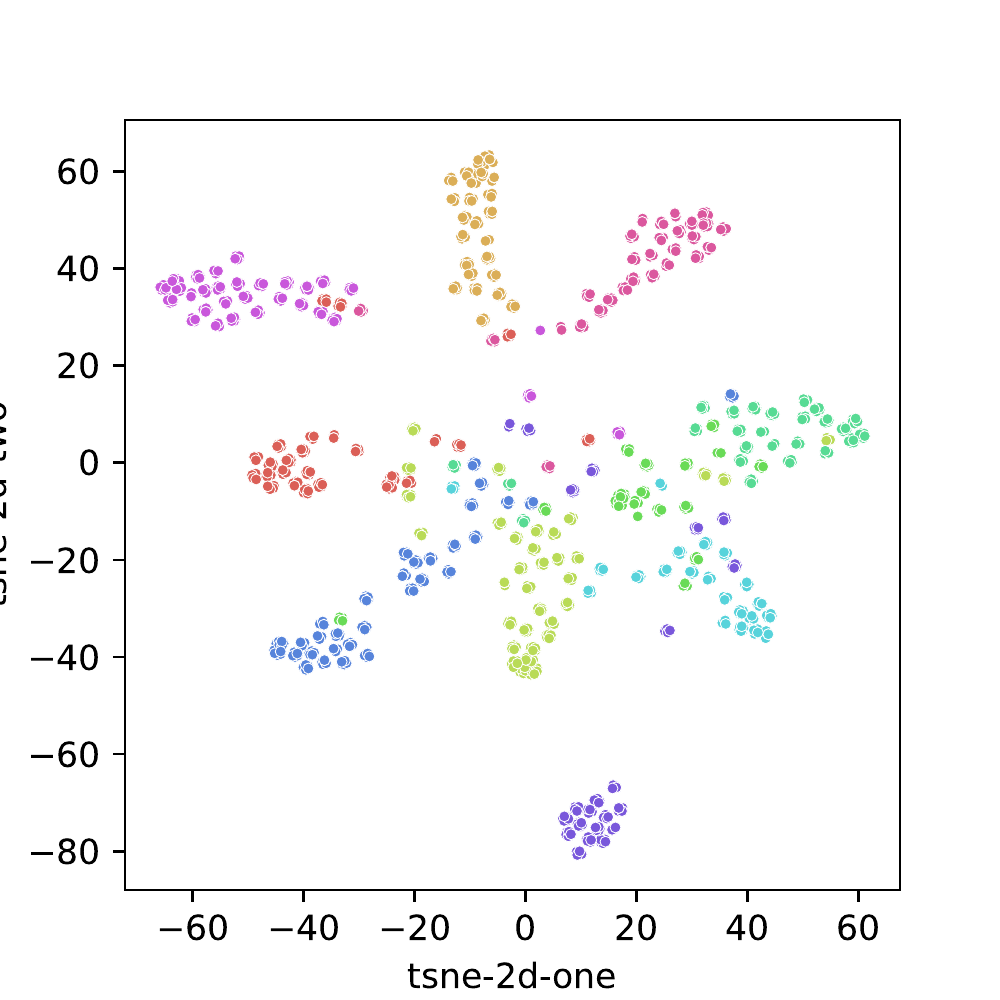}
    \end{minipage}}
    \label{fig:tsne_ir}
\subfigure[RBNN~\cite{lin2020rotated}]{
    \begin{minipage}{0.22\textwidth}
    \includegraphics[width=\textwidth]{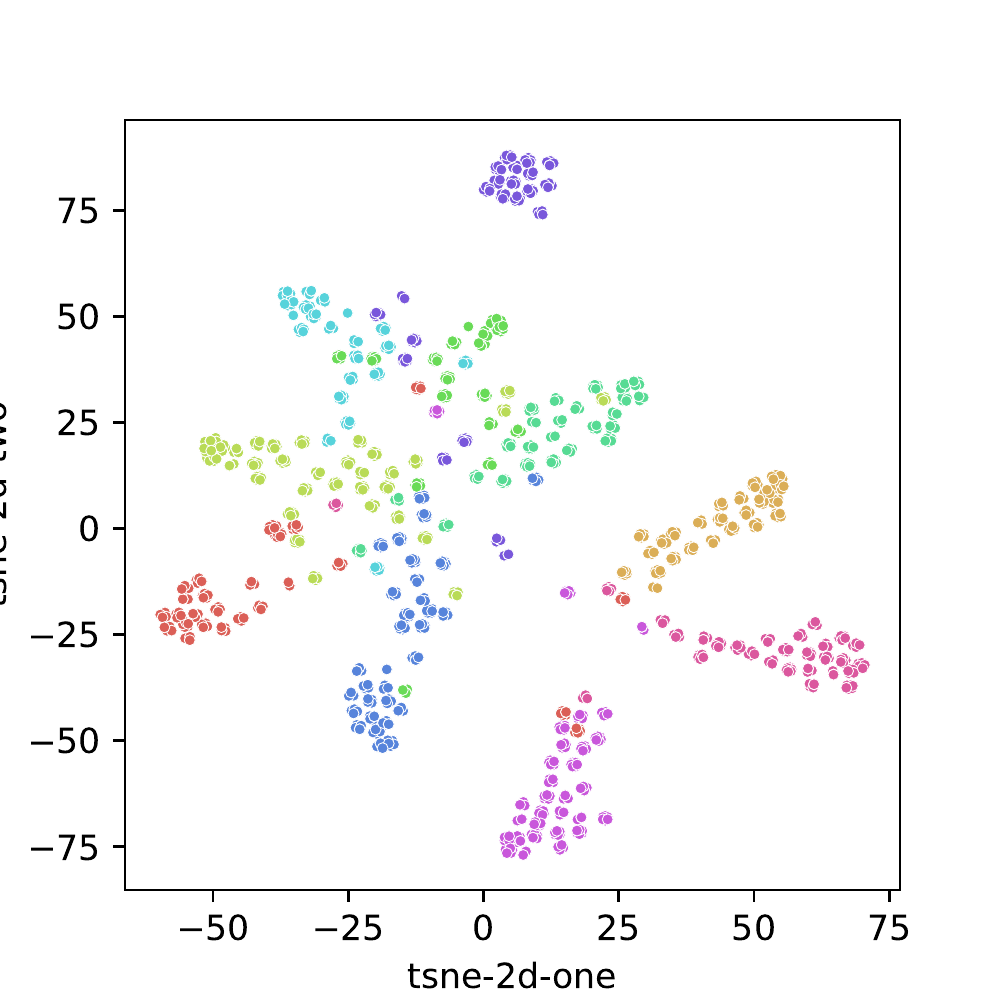}
    \end{minipage}}
    \label{fig:tsne_rbnn}
\subfigure[\emph{CMIM} (ours)]{
    \begin{minipage}{0.22\textwidth}
    \includegraphics[width=\textwidth]{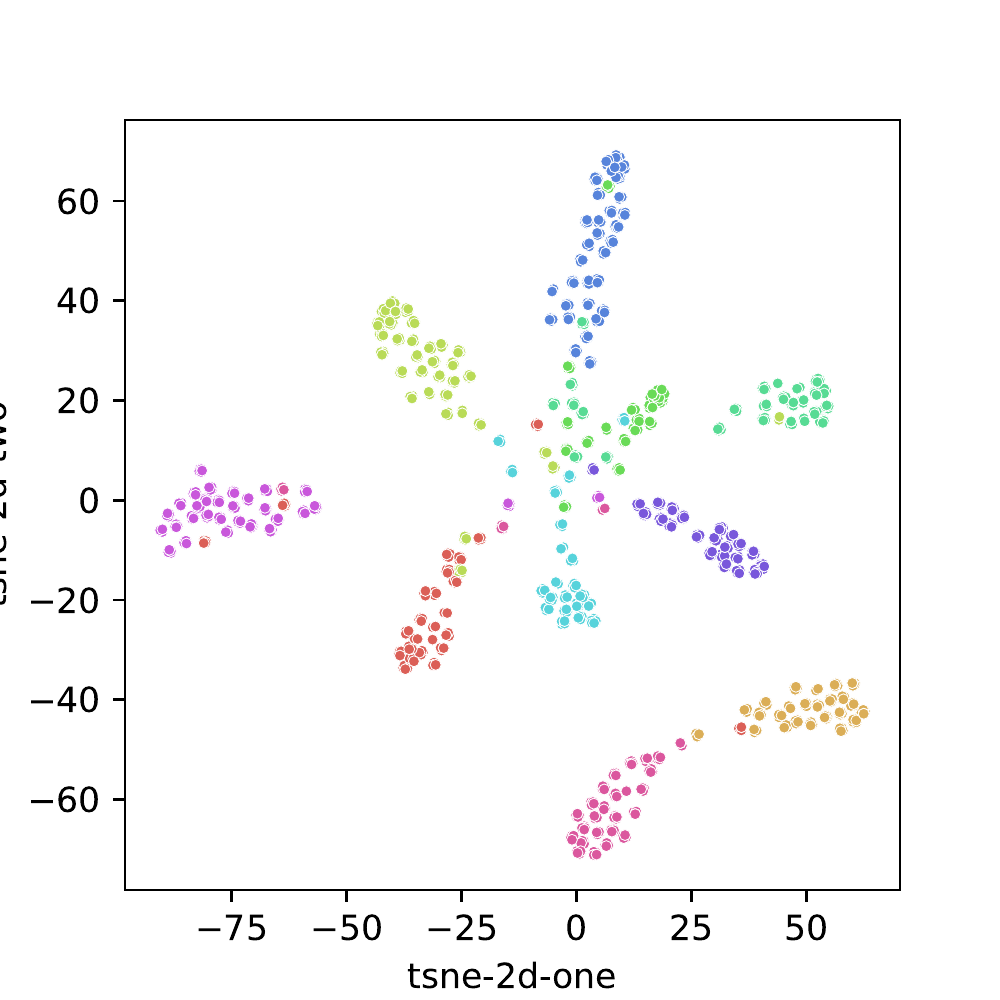}
    \end{minipage}}
    \label{fig:tsne_mim}
\caption{t-SNE~\cite{van2008visualizing} visualization of the activations representing for random 10 classes in CIFAR-100. Every color represents a different class. We can clearly witness the improvement of our method for learning better binary representations.}
\label{fig:tsne}
\vspace{-0.8em}
\end{figure}

Besides the theoretical formulation from the perspective of mutual information maximization, we also provide an intuitive explanation about \emph{CMIM}. As illustrated in Fig.~\ref{fig:intuition}, we strengthen the representation ability of binary activations (see Fig.~\ref{fig:tsne}) via designing a proxy task under the contrastive learning framework. By embedding the activations to the contrastive space and pull-and-push the paired embeddings, the BNNs can learn better representations from this difficult yet effective auxiliary contrastive learning task. Note that even though we only pick up two images to formulate Fig.~\ref{fig:intuition}, the actual number of negative samples can be huge in practice (\textit{e.g.} 16,384 for training ResNet-18 on ImageNet), benefit from the MemoryBank~\cite{wu2018unsupervised} technique.

With this property, we speculate that the contrastive pairing works as the data augmentation, which contributes to our method. This additional pairing provides more information for training the BNNs, thus \emph{CMIM} can be treated as an overfitting-mitigated module. We also conduct experiments in the Section~\ref{sec:n_nce} and \ref{sec:regularization} to validate our speculation. 


\noindent\textbf{Comparison with other contrastive learning methods.} The key idea of contrastive learning is to pull representations close in positive pairs and push representations apart in negative pairs in a contrastive space. Several self-supervised learning methods are rooted in well-established idea of the mutual information maximization, such as Deep InfoMax~\cite{hjelm2018learning}, Contrastive Predictive Coding~\cite{oord2018representation}, MemoryBank~\cite{wu2018unsupervised}, Augmented Multiscale DIM~\cite{bachman2019learning}, MoCo~\cite{he2020momentum} and SimSaim~\cite{chen2021exploring}. These are based on NCE~\cite{gutmann2010noise} and InfoNCE~\cite{hjelm2018learning} which can be seen as a lower bound on mutual information~\cite{poole2019variational}. In the meantime, Tian~\textit{et. al.}~\cite{tian2019contrastive} and Chen~\textit{et. al.}~\cite{chen2021wasserstein} generalize the contrastive idea into the content of knowledge distillation (KD) to pull-and-push the representations of teacher and student. 

Our formulation of \emph{CMIM}-BNN absorbs the core idea (\textit{i.e.} construct the appropriate positive and negative pairs for contrastive loss) of the existing contrastive learning methods, especially the contrastive knowledge distillation methods, CRD~\cite{tian2019contrastive} and WCoRD~\cite{chen2021wasserstein}. However, our approach has several differences from those methods. 
Firstly, our work can not be treated as a simply application with the teacher-and-student framework. In KD, the teacher is basically fixed to offer additional supervision signals and is not optimizable. But in our formulation, we leverage the exclusive structure of BNN, where FP and binary activations exist in the same forward pass, \textit{i.e.} only one BNN is involved, without using another network as a teacher. Therefore, the accuracy improvement of the BNN trained by our method is purely benefited from the activation alignment in a contrastive way, rather than a more accurate teacher network. 
Secondly, due to the particular structure of BNNs (Eq.~\ref{eq:critic_1}), our critic function is largely different from the normal critic in contrastive learning (see Eq.~\ref{eq:critic_1} and Eq.~\ref{eq:critic_new}). Importantly, the critic functions of CRD and WCoRD must utilize a fully-connected layer over the representations to transform them into the same dimension and further normalize them by $L_2$ norm before the inner product, but ours does not. 
In the literature of binarization, our designed critic function act as an activation flip as we discussed below Eq.~\ref{eq:critic_new}. 
Thirdly, instead of only using the activation of the final layer, we align the activations layer-by-layer with a hyperparameter to adjust the weight of each layer as shown in Eq.~\ref{eq:loss2}, which is a more suitable design for BNN. 
In conclusion, using contrastive objective as a tool to realize mutual information maximization for our network binarization is new. 
\section{Experiments}
In this section, we first conduct experiments to compare with existing state-of-the-art methods in image classification. Following popular settings in most studies, we use CIFAR-10/100~\cite{krizhevsky2009learning} and ImageNet ILSVRC-2012~\cite{deng2009imagenet} to validate the effectiveness of our proposed binarization method. Besides comparing our method with the SoTA methods, we design experiments in semantic segmentation and depth estimation tasks on the NYUD-v2~\cite{silberman2012indoor} dataset to testify the generalization ability of our method. Meanwhile, we conduct a series of ablation studies to verify the effectiveness of our proposed technique, and we empirically explain the efficacy of \emph{CMIM} from the perspective of mitigating overfitting. All experiments are implemented using PyTorch \cite{paszke2019pytorch} with one NVIDIA RTX 6000 while training on CIFAR-10/100 and NYUD-v2, and four GPUs on ImageNet.

\noindent \textbf{Experimental Setup.} On CIFAR-10/100, the BNNs are trained by \emph{CMIM} for 400 epochs with batch size of 256, initial learning rate of 0.1 and cosine learning rate scheduler. We adopt SGD optimizer with momentum of 0.9 and weight decay of 1e-4. On ImageNet, binary models are trained for 100 epochs with batch size of 256. SGD optimizer is applied with momentum of 0.9, weight decay of 1e-4, initial learning rate of 0.1 with cosine learning rate scheduler (for fair comparison, we also use ADAM optimizer in some ResNet-variant settings).

\subsection{Experimental Results}

\begin{table*}[!t]
\begin{minipage}{.49\textwidth}
    \centering
    \caption{Top-1 accuracy (\%) on CIFAR-10 (C-10) and CIFAR-100 (C-100) test set. The higher the better. W/A denotes the bit number of weights/activations. }
    \scalebox{0.7}{
    \begin{tabular}{ccccc}
        \toprule
        \multirow{2}{*}{Topology}  & \multirow{2}{*}{Method}    & Bit-width       & Acc.(\%)  & Acc.(\%)     \\
                             &                & (W/A)           & (C-10)  & (C-100)      \\ \midrule
                             & Full-precision       & 32/32           & 92.1    &   70.7   \\
                             & DoReFa~\cite{zhou2016dorefa}  & 1/1             & 79.3     & -     \\
                             & QSQ~\cite{gong2019differentiable}  & 1/1         & 84.1     & -     \\
                             & SLB~\cite{yang2020searching}  & 1/1         & 85.5     & -     \\
                             & LNS~\cite{han2020training}  & 1/1         & 85.8     & -     \\
      ResNet                & IR-Net~\cite{qin2020forward}  & 1/1             & 86.5   & 65.6       \\
        -20                  & RBNN~\cite{lin2020rotated}   & 1/1             & 87.0    & 66.0      \\
                             & IR-Net + \emph{CMIM}    & 1/1             & \textbf{87.3}  & \textbf{68.1}     \\ 
                             & RBNN + \emph{CMIM}          & 1/1             & \textbf{87.6}  & \textbf{68.2}     \\ \hline
                             & Full-precision       & 32/32           & 93.0    &   72.5   \\
                             & RAD~\cite{ding2019regularizing}  & 1/1             & 90.5     & -     \\
                             & Proxy-BNN~\cite{he2020proxybnn}   & 1/1             & 91.8    & 67.2      \\
      ResNet                & IR-Net~\cite{qin2020forward}  & 1/1             & 91.6   & 64.5       \\
        -18                  & RBNN~\cite{lin2020rotated}  & 1/1             & 92.2   & 65.3      \\
                             & IR-Net + \emph{CMIM}   & 1/1             & \textbf{92.2}  & \textbf{71.2}     \\ 
                             & RBNN + \emph{CMIM}      & 1/1             & \textbf{92.8}  & \textbf{71.4}     \\ \hline
                             & Full-precision       & 32/32           & 94.1    &   73.0   \\
                             & XNOR~\cite{rastegari2016xnor}  & 1/1             & 90.5     & -     \\
                             & DoReFa~\cite{zhou2016dorefa}  & 1/1             & 90.2     & -     \\
                             & RAD~\cite{ding2019regularizing}  & 1/1             & 90.5     & -     \\
         VGG                 & QSQ~\cite{gong2019differentiable}  & 1/1          & 90.0     & -     \\
        -small               & SLB~\cite{yang2020searching}  & 1/1         & 92.0      & -     \\
                             & Proxy-BNN~\cite{he2020proxybnn}   & 1/1             & 91.8    & 67.2      \\
                             & IR-Net~\cite{qin2020forward}  & 1/1                & 90.4   & 67.0       \\
                             & RBNN~\cite{lin2020rotated}  & 1/1                  & 91.3   & 67.4       \\
                             & IR-Net + \emph{CMIM}   & 1/1             & \textbf{92.0}  & \textbf{70.0}     \\
                             & RBNN + \emph{CMIM}    & 1/1             & \textbf{92.2}  & \textbf{71.0}     \\
                             \bottomrule
    \end{tabular}}
    \label{tabel:cifar}
\end{minipage}
\hfill
\begin{minipage}{.49\textwidth}
    \centering
    \caption{Top-1 and Top-5 accuracy on ImageNet. ${\dagger}$ represents the architecture which varies from the standard ResNet architecture but in the same FLOPs level.}
    \scalebox{0.71}{
    \begin{tabular}{ccccc}\toprule
            \multirow{2}{*}{Topology} & \multirow{2}{*}{Method}  & BW & Top-1 & Top-5 \\ 
                                      &                          & (W/A)     & (\%)  & (\%) \\\midrule
                      & Full-precision & 32/32           & 69.6       & 89.2       \\
                      & ABC-Net~\cite{lin2017towards} & 1/1             & 42.7       & 67.6       \\
                      & XNOR-Net~\cite{rastegari2016xnor}    & 1/1             & 51.2       & 73.2       \\
                      & BNN+~\cite{hubara2016binarized}   & 1/1             & 53.0       & 72.6       \\
\multirow{2}*{ResNet-18}   & DoReFa~\cite{zhou2016dorefa}   & 1/2             & 53.4       & -          \\
                      & XNOR++~\cite{bulat2019xnor}    & 1/1             & 57.1       & 79.9       \\
                      & BiReal~\cite{liu2020bi}  & 1/1            & 56.4       & 79.5       \\
                      & IR-Net~\cite{qin2020forward}   & 1/1             & 58.1       & 80.0       \\
                      & RBNN~\cite{lin2020rotated}  & 1/1             & 59.9       & 81.0       \\
                      & BiReal + CMIM     & 1/1             & \textbf{60.1} & \textbf{81.3} \\ 
                      & IR-Net + CMIM     & 1/1             & \textbf{61.2} & \textbf{83.0} \\ 
                      & RBNN + CMIM   & 1/1             & \textbf{62.5}       & \textbf{84.2}       \\\cline{2-5}
                      & {ReActNet~\cite{liu2020reactnet}}$^{\dagger}$   & 1/1             & 69.4       & 85.5       \\
                      & {ReActNet + CMIM}$^{\dagger}$     & 1/1             & \textbf{71.0} & \textbf{86.3} \\ 
                      \midrule
                      & Full-precision & 32/32           & 73.3       & 91.3       \\
                      & ABC-Net~\cite{lin2017towards}  & 1/1             & 52.4       & 76.5       \\
                      & XNOR-Net~\cite{rastegari2016xnor}    & 1/1             & 53.1       & 76.2       \\
                      & BiReal~\cite{liu2020bi}   & 1/1            & 62.2       & 83.9       \\
\multirow{2}*{ResNet-34} & XNOR++~\cite{bulat2019xnor}   & 1/1             & 57.1       & 79.9       \\
                      & IR-Net~\cite{qin2020forward}    & 1/1             & 62.9       & 84.1       \\
                      & LNS~\cite{han2020training}    & 1/1             & 59.4       & 81.7       \\
                      & RBNN~\cite{lin2020rotated}   & 1/1             & 63.1       & 84.4       \\
                      & IR-Net + CMIM     & 1/1             & \textbf{64.9}   & \textbf{85.8} \\ 
                      & RBNN + CMIM   & 1/1                 & \textbf{65.0}   & \textbf{85.7}       \\\bottomrule
        \end{tabular}}
    \label{tabel:imagenet}
\end{minipage}
\vspace{-0.23in}
\end{table*}

\noindent \textbf{CIFAR-10/100} are widely-used  image  classification datasets, where each consists of 50K training images and 10K testing images of size 32×32 divided into 10/100 classes. 10K training images are randomly sampled for cross-validation and the rest images are utilized for training. Data augmentation strategy includes random crop and random flipping as in~\cite{he2016deep} during training.

For ResNet-20, we compare with DoReFa~\cite{zhou2016dorefa}, QSQ~\cite{gong2019differentiable}, SLB~\cite{yang2020searching}, LNS~\cite{han2020training}, IR-Net~\cite{qin2020forward} and RBNN~\cite{lin2020rotated}. For ResNet-18, RAD~\cite{ding2019regularizing}, Proxy-BNN~\cite{he2020proxybnn}, IR-Net and RBNN are chosen to be the benchmarks. For VGG-small, our method is compared with IR-Net and RBNN, \textit{etc}. 

As presented in Table~\ref{tabel:cifar}, \emph{CMIM} constantly outperforms other SOTA methods. On CIFAR-100, our method achieves 2.5\%, 6.1\% and 4.0\% performance improvement with ResNet-20, ResNet-18 and VGG-small architectures, respectively. To show the pile-up property, we add \emph{CMIM} on different baseline methods, and we can obviously observe the accuracy gain with \emph{CMIM}.

\noindent \textbf{ImageNet} is a dataset with 1.2 million training images and 50k validation images equally divided into 1K classes. ImageNet has greater diversity, and its image size is 469×387 (average). We report the single-crop evaluation result using 224×224 center crop from images. 

For ResNet-18, we compare our method with XNOR-Net~\cite{rastegari2016xnor}, ABC-Net~\cite{lin2017towards}, DoReFa~\cite{zhou2016dorefa}, BiReal~\cite{liu2020bi}, XNOR++~\cite{bulat2019xnor}, IR-Net~\cite{qin2020forward}, RBNN~\cite{lin2020rotated}. For ResNet-34, we compare our method with BiReal, IR-Net and RBNN, \textit{etc}. All experimental results are either taken from their published papers or reproduced by ourselves using their code. As demonstrated in Table~\ref{tabel:imagenet}, our proposed method exceeds all the methods in both top-1 and top-5 accuracy. Particularly, \emph{CMIM} achieves around 1.3\% Top-1 accuracy gain with ResNet-18 architecture, as well as 1.9\% Top-1 accuracy improvement with ResNet-34 architecture, compared with the SoTA RBNN method.

\subsection{Number of Negative Samples in \emph{CMIM}}
\label{sec:n_nce}

\begin{figure}[!t]
\subfigure[Full network]{
    \begin{minipage}{0.22\textwidth}
    \includegraphics[width=\textwidth]{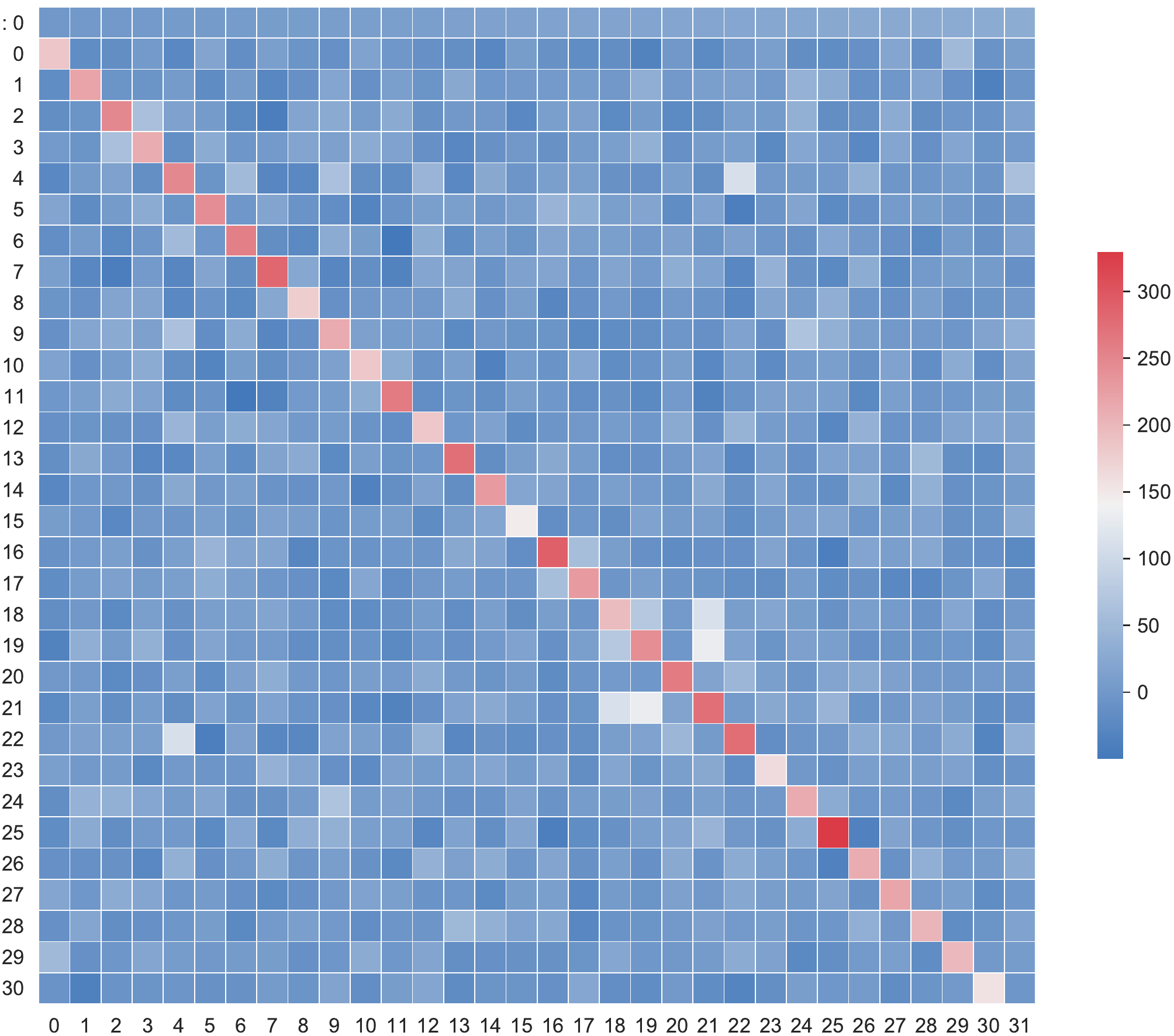}
    \end{minipage}
    \label{fig:corr_full}}
\subfigure[IR-Net~\cite{qin2020forward}]{
    \begin{minipage}{0.22\textwidth}
    \includegraphics[width=\textwidth]{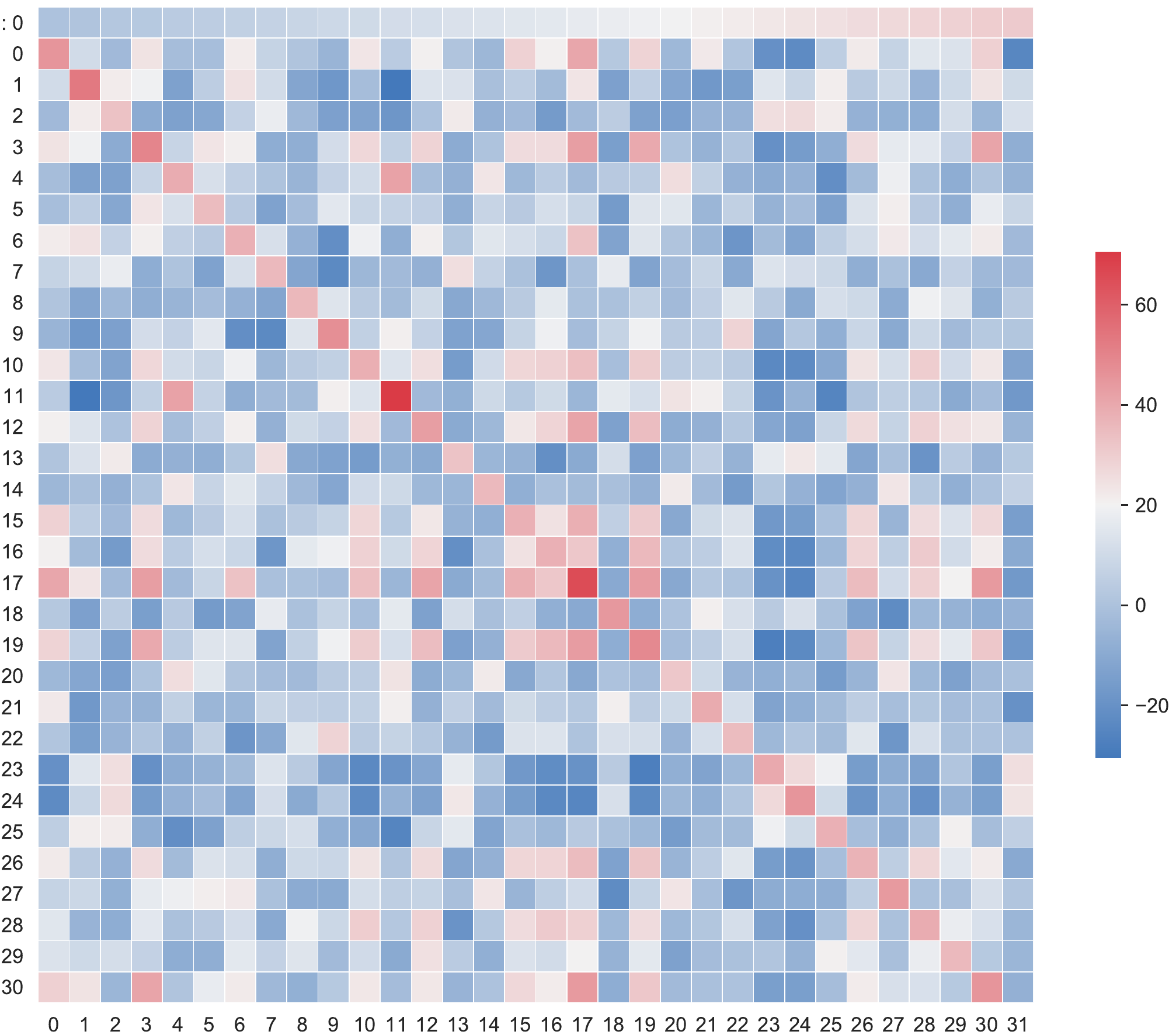}
    \end{minipage}
    \label{fig:corr_ir}}
\subfigure[RBNN~\cite{lin2020rotated}]{
    \begin{minipage}{0.22\textwidth}
    \includegraphics[width=\textwidth]{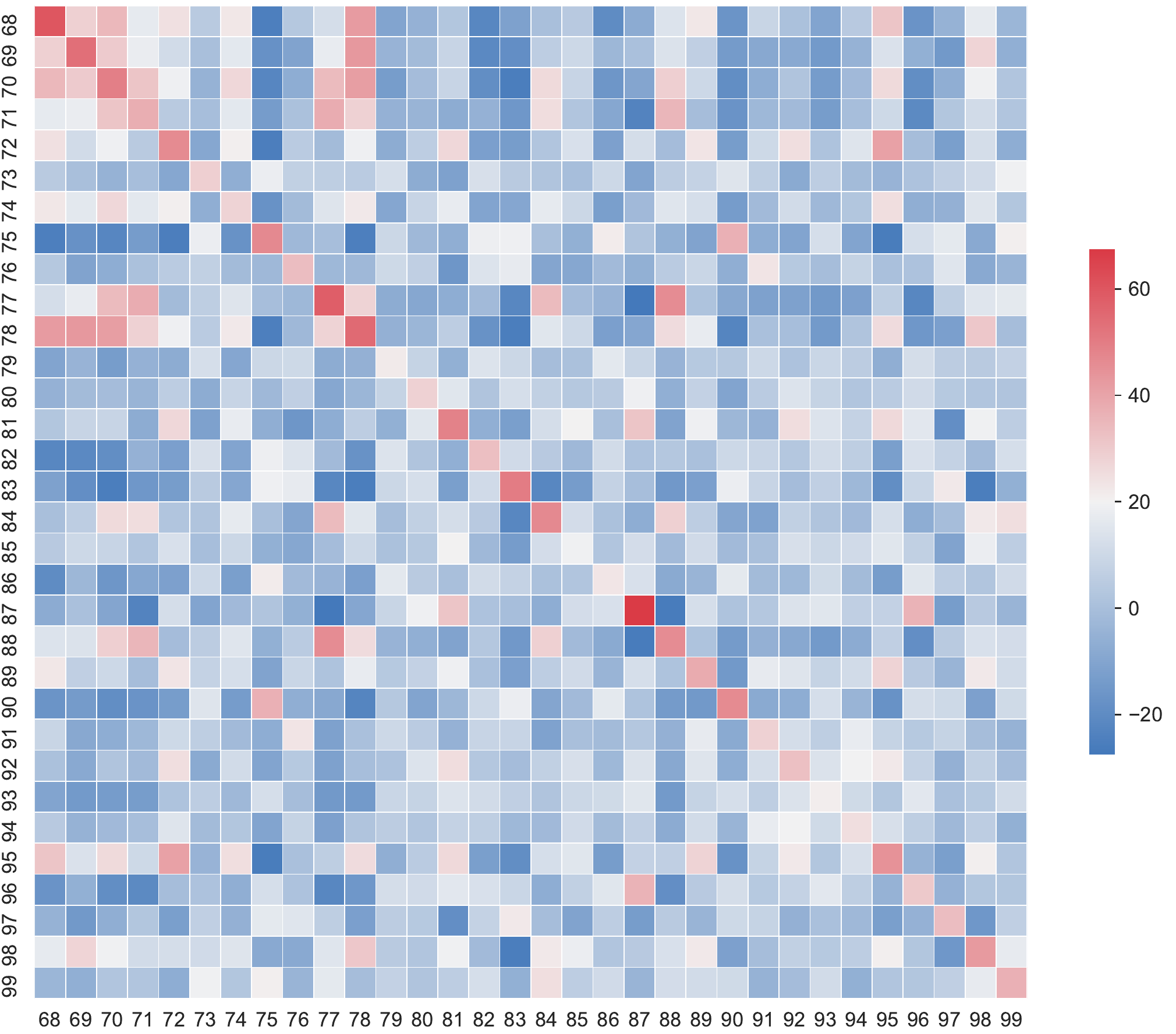}
    \end{minipage}
    \label{fig:corr_rbnn}}
\subfigure[\emph{CMIM} (ours)]{
    \begin{minipage}{0.22\textwidth}
    \includegraphics[width=\textwidth]{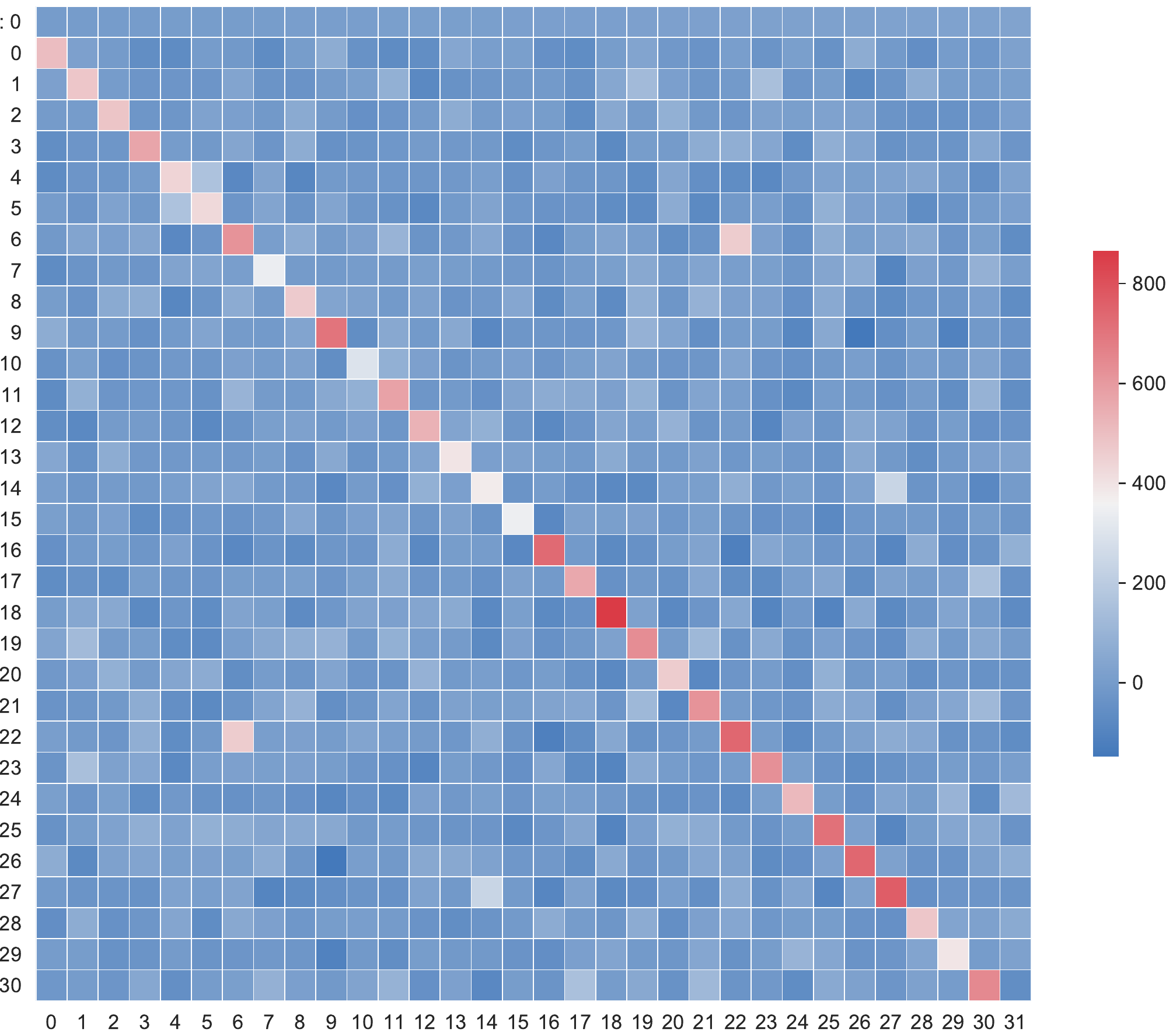}
    \end{minipage}
    \label{fig:corr_mim}}
\label{fig:correlation}
\subfigure[n\_nce on CIFAR-10]{
    \begin{minipage}{0.32\textwidth}
    \includegraphics[width=\textwidth]{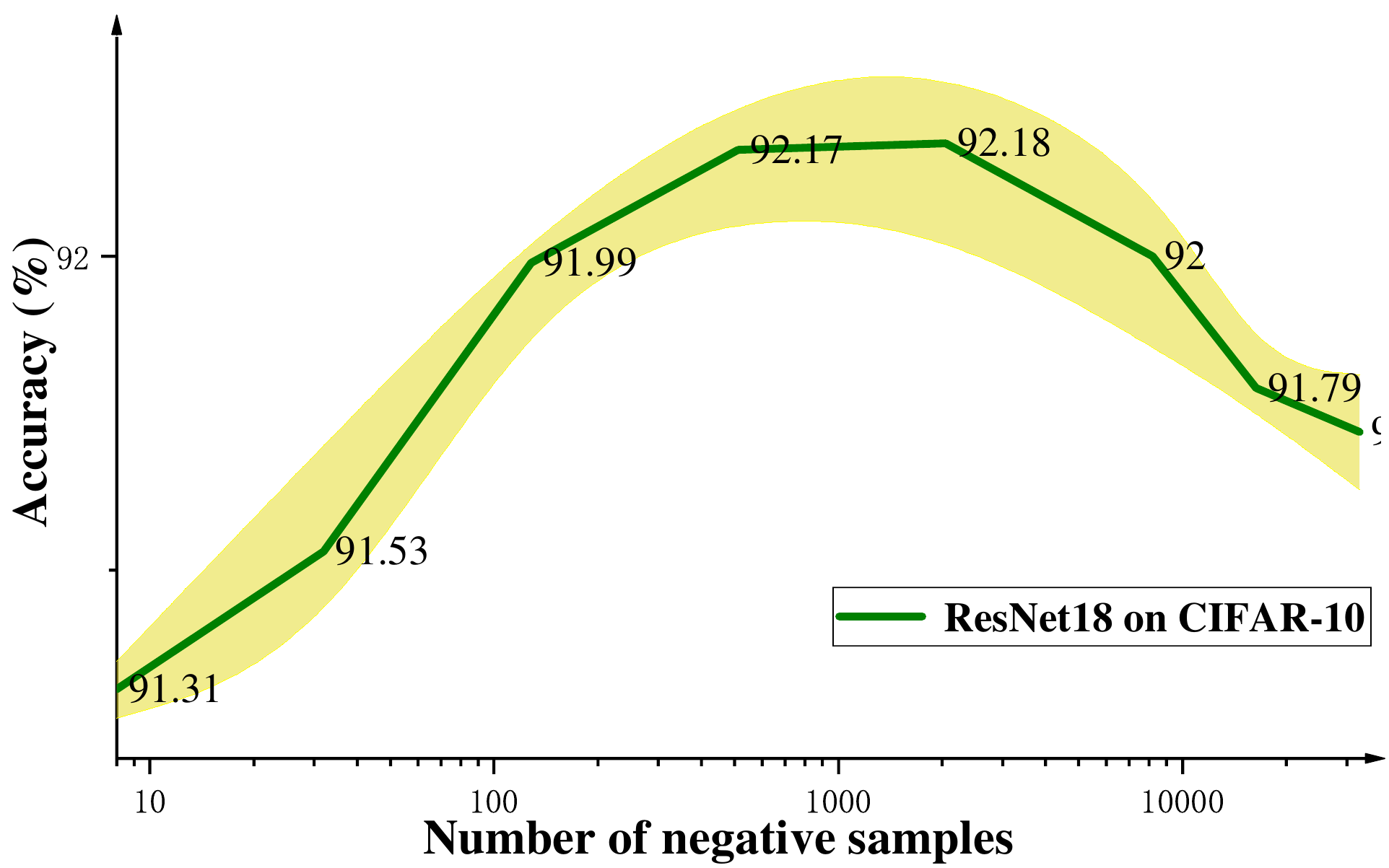}
    \end{minipage}
    \label{fig:n_nce_10}}
\subfigure[n\_nce on CIFAR-100]{
    \begin{minipage}{0.32\textwidth}
    \includegraphics[width=\textwidth]{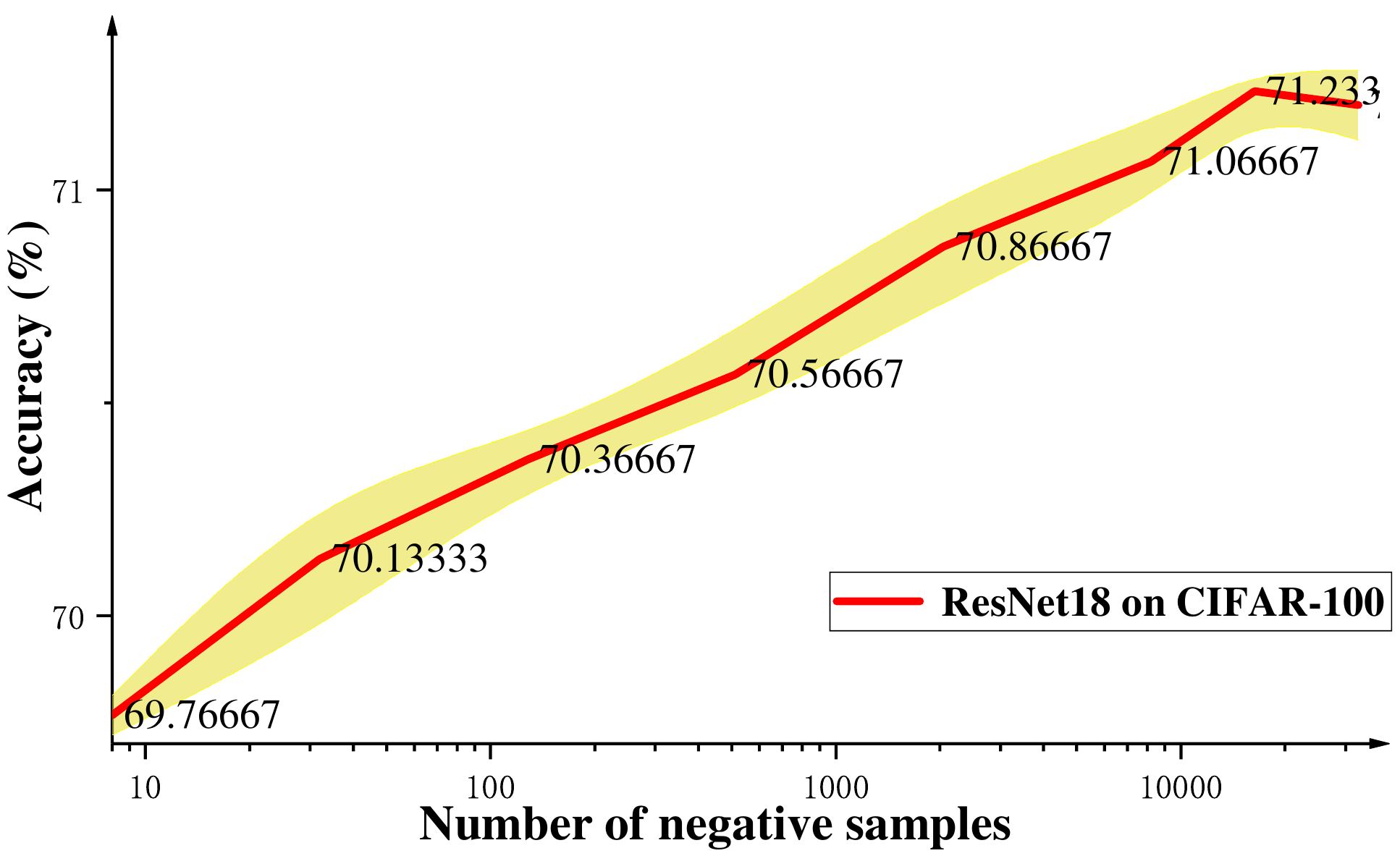}
    \end{minipage}
     \label{fig:n_nce_100}}
\subfigure[RBNN~\cite{lin2020rotated}]{
    \begin{minipage}{0.28\textwidth}
    \includegraphics[width=0.9\textwidth]{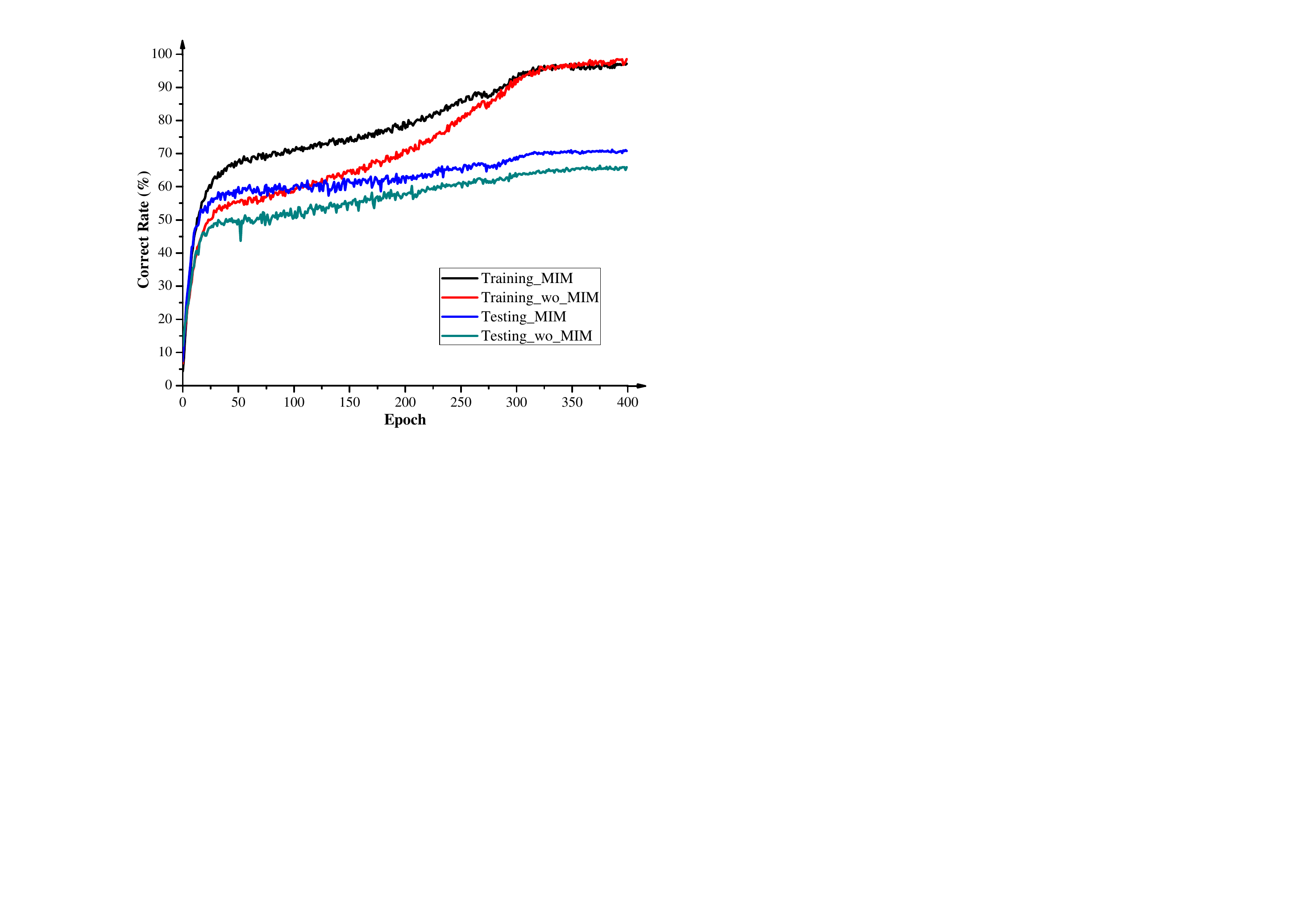}
    \end{minipage}
    \label{fig:regularization}}
\caption{In-depth analysis on different aspects of the proposed approach including correlation maps (a-d) the effect of number of negative samples in contrastive mutual information maximization (e,f), and training and testing curves (g).}
\label{fig:further_analysis}
\end{figure}

The number of negative samples n\_nce is an important hyper-parameter in our method, which ensures the estimation accuracy level of the optimized distribution in Eq.~\ref{eq:ctl4}. We perform experiments with ResNet18 on CIFAR-100 for parameter analysis of n\_nce, with range from $2^0$ to $2^{15}$. As the results in Fig.~\ref{fig:n_nce_100} and ~\ref{fig:n_nce_10} presented, the accuracy arises with increasing n\_nce, which also validates our speculation in the Section~\ref{sec:method} that the contrastive pairing module, serving as a data augmentation module in training, contributes to the performance improvement of \emph{CMIM}. 

\subsection{Mitigate Overfitting}
\label{sec:regularization}

A good training objective should consistently improve the model performance in testing set~\cite{wu2018unsupervised}. We investigate the relation between the training and the testing performance \textit{w.r.t.} training iterations. Fig.~\ref{fig:regularization} shows that (1) the binary ResNet-18 can reach 100\% on training set of CIFAR-100, which means its representative ability is enough for this dataset; (2) the testing performance of the BNN trained with \emph{CMIM} loss is much better on the final stage, while the training performance is relatively lower. This is a clear sign of mitigating overfitting. In addition, as the results shown in the Table~\ref{table:ablation100}, we can observe the phenomenon that the accuracy gain on CIFAR-100 is more noticeable than the gain on ImageNet. This phenomenon can also be explained from the perspective of mitigating overfitting. Since the contrastive pairing (data augmentation for the proxy contrastive learning task) plays a significant role in improving the performance of BNNs, and the data for training is sufficient on ImageNet than on CIFAR. The overfitting issue is not that severe on ImageNet. Hence, our binarization method could be more suitable for relatively data-deficient tasks.

\subsection{Ablation study}

We conduct a series of ablative studies of our proposed method in CIFAR-10/100 and ImageNet datasets with the ResNet18 architecture. By adjusting the coefficient $\lambda$ in the loss function $\mathcal{L}_{\emph{CMIM}}$ (Eq. \ref{eq:loss2}), where $\lambda = 0$ equals to no \emph{CMIM} loss are added as our baseline. In the ablative studies, we introduce IR-Net~\cite{qin2020forward} as our baseline on all the datasets. The results are shown in Table \ref{table:ablation100}. With $\lambda$ increasing, the improving performance validates the efficacy of \emph{CMIM} loss.

\begin{table}
\centering
\caption{Ablation study of \emph{CMIM}. The results are presented in the form of accuracy rate (\%). $\lambda = 0$ denotes no \emph{CMIM} loss added, serving as our baseline.}
\scalebox{0.8}{
    \begin{tabular}[width=0.9\textwidth]{c|cccccccc}
    \toprule
    \diagbox{Dataset}{$\lambda$}         & 0 (baseline)    & 0.2   & 0.4   & 0.8   & 1.6              & 3.2      & 6.4           & 12.8 \\ 
    \specialrule{0.8pt}{0pt}{0pt}
    CIFAR-10                & 87.59  & 90.92 & 91.63 & 92.06 & \textbf{92.18}   & 91.89    & 91.32         & 91.01\\
    CIFAR-100               & 64.53  & 68.21 & 69.31 & 70.67 & 70.86            & 71.09    &\textbf{71.19} &71.17\\
    \midrule
    ImageNet-1K               & 58.03  & 59.29 & 59.99 & \textbf{61.22} & 61.17            & 61.02    & 60.64 & 59.7\\
    \bottomrule
    \end{tabular}}
    \label{table:ablation100}
\end{table}

\subsection{Generalization Ability}
\label{sec:generalization}
To study the dependence of the binary activations from the same layer, we visualize the correlation matrix of those activations by using the shade of the color to represent the cosine similarity of two activations. Red stands for two activations are similar and blue \textit{vice versa}. As shown in Figs.~\ref{fig:corr_full}-\ref{fig:corr_mim}, \emph{CMIM} captures more intra-class correlations (diagonal boxes are redder) and alleviates more inter-class correlations (non-diagonal boxes are bluer). Those intensified representative activations are constructive for fine-tuning down-stream tasks. 

\begin{figure}[!t]
\subfigure[Visualized results]{
    \begin{minipage}{0.64\textwidth}
    \includegraphics[width=\textwidth]{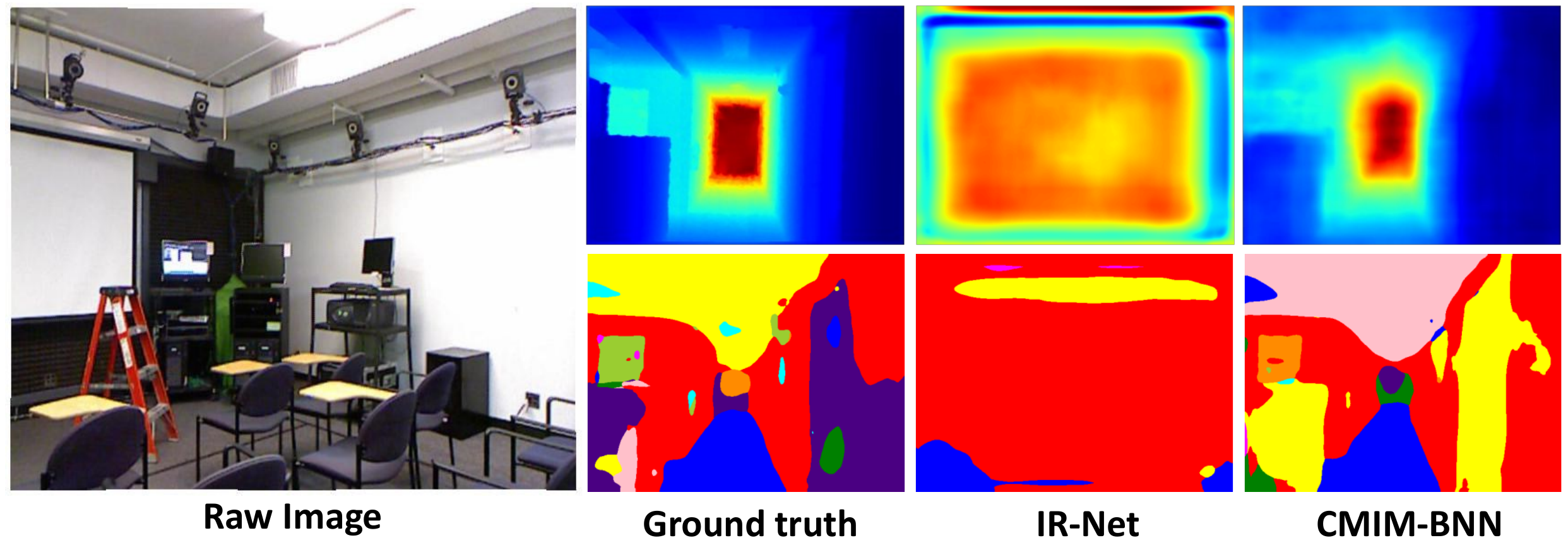}
    \end{minipage}
    \label{fig:nyud}}
\subfigure[Quantitative results]{
    \begin{minipage}{0.3\textwidth}
    \includegraphics[width=0.9\textwidth]{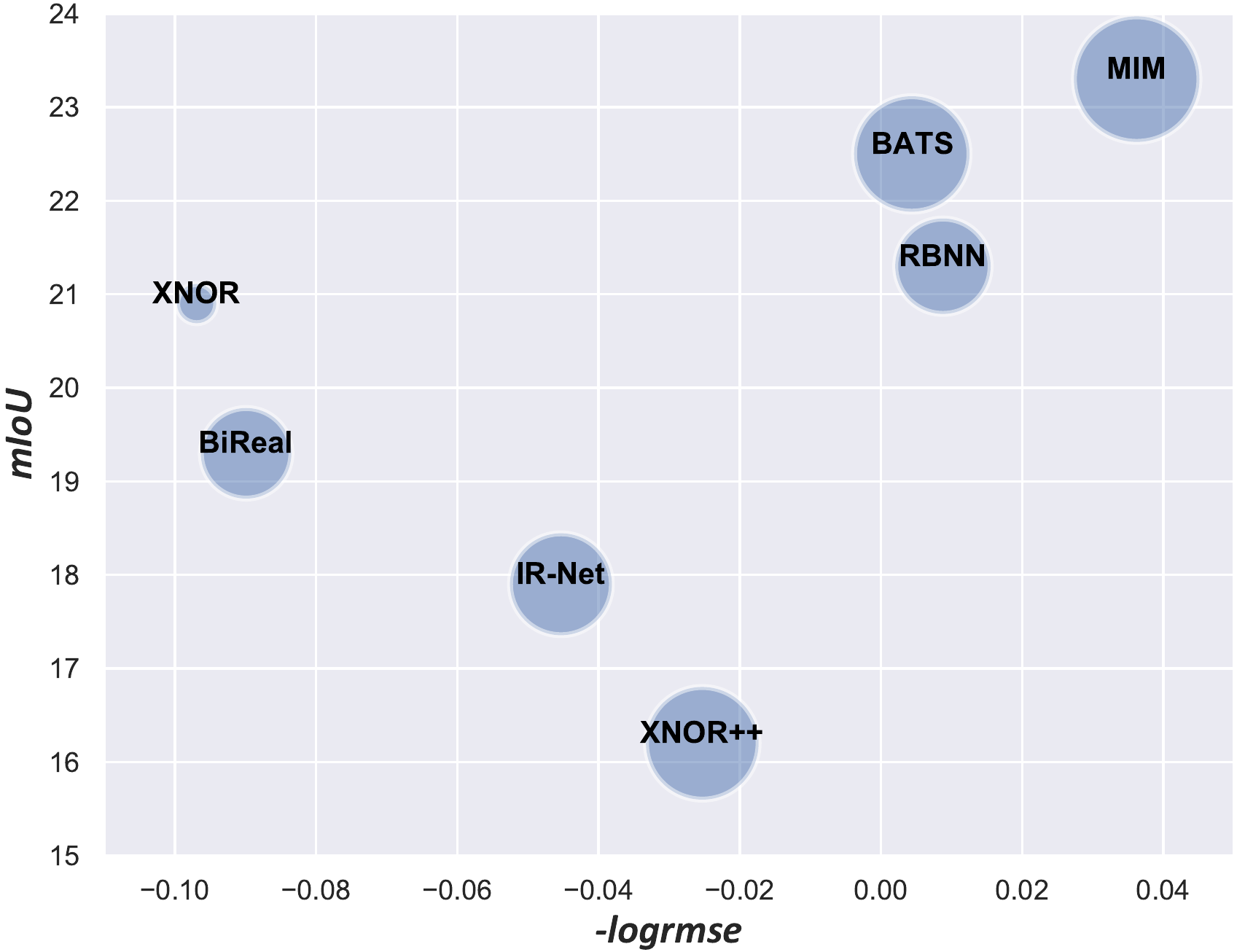}
    \end{minipage}
    \label{fig:generalization}}
\caption{Results of depth estimation and segmentation on NYUD-v2}
\label{fig:further_analysis}
\end{figure}

To further evaluate the generalization capacity of the learned binary features, we transfer the learned binary backbone to the image segmentation and depth estimation on NYUD-v2 dataset. We follow the standard pipeline for fine-tuning. A prevalent practice is to pre-train the backbone network on ImageNet and fine-tune it for the downstream tasks. Thus, we conduct experiments with DeepLab heads with the binary ResNet18 backbone. While fine-tuning, the learning rate is initialized to 0.001 and scaled down by 10 times after every 10K iterations and we fix the binary backbone weights, only updating the task-specific heads layers. The results are presented in Fig.~\ref{fig:generalization},  X-axis is the depth estimation accuracy (-logrmse, higher is better), Y-axis is segmentation performance (mIoU, higher is better) and the size of dot denotes the performance of classification (bigger is better). The visualization results are presented in Fig.~\ref{fig:nyud}. We can observe that the models with backbone pre-trained by \emph{CMIM} outperform other methods on both segmentation and depth estimation tasks.

\section{Conclusion}
\label{sec:con}
In this paper, we investigate the activations of BNNs by introducing mutual information to measure the distributional similarity between the binary and full-precision activations. We establish a proxy task via contrastive learning to maximize the targeted mutual information between those binary and real-valued activations. We name our method \emph{CMIM}-BNN. Because of the push-and-pull scheme in the contrastive learning, the BNNs optimized by our method have better representation ability, benefiting downstream tasks, such as classification and segmentation, \textit{etc}. We conduct experiments on CIFAR, ImageNet (for classification) and NYUD-v2 (fine-tuning for depth estimation and segmentation). The results show that CMIM outperforms several state-of-the-art binarization methods on those tasks. 

\noindent \textbf{Acknowledgements.} This research was partially supported by NSF CNS-1908658 (ZZ,YY), NeTS-2109982 (YY), Early Career Scheme of the Research Grants Council (RGC) of the Hong Kong SAR under grant No. 26202321 (DX), HKUST Startup Fund No. R9253 (DX) and the gift donation from Cisco (YY). This article solely reflects the opinions and conclusions of its authors and not the funding agents.

\clearpage
%
%
\bibliographystyle{splncs04}
\bibliography{egbib}

\end{document}